%% file: neurips_2025.tex
\documentclass{article}

\usepackage[final]{neurips_2025}

\usepackage[utf8]{inputenc} 
\usepackage[T1]{fontenc}    
\usepackage{hyperref}       
\usepackage{url}            
\usepackage{booktabs}       
\usepackage{amsfonts}       
\usepackage{nicefrac}       
\usepackage{microtype}      
\usepackage{xcolor}         

\usepackage{adjustbox}

\usepackage{colortbl}
\usepackage{pifont}

\usepackage{caption}
\usepackage{subcaption}

\usepackage{titletoc}

\definecolor{indigo}{rgb}{0.0, 0.25, 0.42}

\usepackage{booktabs, multirow} 
\usepackage[utf8]{inputenc} 
\usepackage[T1]{fontenc}    
\usepackage[]{hyperref}       
\usepackage{url}            
\usepackage{booktabs}       
\usepackage{amsfonts}       
\usepackage{nicefrac}       
\usepackage{microtype}      
\usepackage{xcolor}         
\usepackage{soul}
\usepackage{xcolor,colortbl} 
\usepackage{changepage,threeparttable} 
\usepackage{tabularx}
\usepackage{layouts}

\definecolor{rank1}{HTML}{e07a5f}
\definecolor{rank2}{HTML}{eab69f}
\definecolor{rank3}{HTML}{f4f1de}
\definecolor{rank4}{HTML}{FFD43B}
\definecolor{rank5}{HTML}{FF7F00}
\definecolor{rank6}{HTML}{E41A1C}

\usepackage{hyperref}

\usepackage{algorithm}
\usepackage{algpseudocode} 

\usepackage{amsmath}
\usepackage{amssymb}
\usepackage{mathtools}
\usepackage{amsthm}
\usepackage{multirow}
\usepackage{wrapfig}

\usepackage{listings}

\usepackage{pgfplots}

\DeclareMathOperator*{\argmin}{arg\,min}

\usepackage[capitalize,noabbrev]{cleveref}

\usepackage{bm}
\usepackage{bbm}

\usepackage{tabu}

\newcommand{\bxi}{\bm{\xi}}

\usepackage{xcolor}

\newcolumntype{H}{>{\setbox0=\hbox\bgroup}c<{\egroup}@{}}

\theoremstyle{plain}

\newtheorem{theorem}{Theorem}

\newtheorem{lemma}{Lemma}

\newtheorem{proposition}{Proposition}

\usepackage{tcolorbox}
\tcbuselibrary{theorems}

\newcolumntype{R}{>{}r<{}}
\newcolumntype{L}{>{}l<{}}
\newcolumntype{M}{R@{}L}

\definecolor{ourspecialtextcolor}{rgb}{0.528, 0.471, 0.701} 

\usepackage{mystyles}

\title{Rao-Blackwell Gradient Estimators for\\ Equivariant Denoising Diffusion}

\author{
  Vinh Tong$^{1,2}$\thanks{Equal contribution.},
  Trung-Dung Hoang$^{4}$\footnotemark[1],
  Anji Liu$^{5}$,
  Guy Van den Broeck$^{3}$,
  Mathias Niepert$^{1,2}$ \\
  $^{1}$University of Stuttgart,
  $^{2}$IMPRS-IS,
  $^{3}$UCLA,
  $^{4}$University of Bern,
  $^{5}$National University of Singapore\\
  \texttt{vinh.tong@ki.uni-stuttgart.de}
}

\usepackage[hypcap=true]{caption}

\newcommand{\citepcolor}[1]{\textcolor{blue}{\citep{#1}}}
\newcommand{\citetcolor}[1]{\textcolor{blue}{\citet{#1}}}

\pgfplotsset{compat=1.18}

\begin{document}

\maketitle

\begin{abstract}

In domains such as molecular and protein generation, physical systems exhibit inherent symmetries that are critical to model. Two main strategies have emerged for learning invariant distributions: designing equivariant network architectures and using data augmentation to approximate equivariance. While equivariant architectures preserve symmetry by design, they often involve greater complexity and pose optimization challenges. Data augmentation, on the other hand, offers flexibility but may fall short in fully capturing symmetries. Our framework enhances both approaches by reducing training variance and providing a provably lower-variance gradient estimator. We achieve this by interpreting data augmentation as a Monte Carlo estimator of the training gradient and applying Rao–Blackwellization. This leads to more stable optimization, faster convergence, and reduced variance, all while requiring only a single forward and backward pass per sample. We also present a practical implementation of this estimator—incorporating the loss and sampling procedure—through a method we call \textit{Orbit Diffusion}. Theoretically, we guarantee that our loss admits equivariant minimizers. Empirically, Orbit Diffusion achieves state-of-the-art results on GEOM-QM9 for molecular conformation generation, improves crystal structure prediction, and advances text-guided crystal generation on the Perov-5 and MP-20 benchmarks. Additionally, it enhances protein designability in protein structure generation. Code is available at \url{https://github.com/vinhsuhi/Orbit-Diffusion.git}.

\end{abstract}

\input{DELTA_sections/introduction}

\input{DELTA_sections/background}

\input{AI4Science/method_reformulate}

\input{DELTA_sections/experiment}

\input{DELTA_sections/related_work}

\input{DELTA_sections/conclusion}

\clearpage


\bibliography{iclr2025_conference}
\bibliographystyle{iclr2025_conference}

\appendix
\newpage 
\begin{center}
    \hrule 
    \startcontents[sections]\vbox{\vspace{4mm}\sc \LARGE Rao-Blackwell Gradient Estimators for Equivariant Denoising Diffusion \\\sc\small \textbf{Additional Material}} \vspace{5mm} \hrule height .5pt
    \printcontents[sections]{l}{0}{\setcounter{tocdepth}{2}}
\end{center}

\include{DELTA_sections/appendix}
\newpage
\section*{NeurIPS Paper Checklist}

\begin{enumerate}

\item {\bf Claims}
    \item[] Question: Do the main claims made in the abstract and introduction accurately reflect the paper's contributions and scope?
    \item[] Answer: \answerYes{} 
    \item[] Justification: The abstract and introduction clearly state the paper's main claims and contributions, which align with the theoretical and experimental results presented.

\item {\bf Limitations}
    \item[] Question: Does the paper discuss the limitations of the work performed by the authors?
    \item[] Answer: \answerYes{} 
    \item[] Justification: We have discussed about our limitation in the Section~\ref{sec:conclusion}.

\item {\bf Theory assumptions and proofs}
    \item[] Question: For each theoretical result, does the paper provide the full set of assumptions and a complete (and correct) proof?
    \item[] Answer: \answerYes{} 
    \item[] Justification: We provided comprehensive proof of our theoretical results in the Appendix section.

    \item {\bf Experimental result reproducibility}
    \item[] Question: Does the paper fully disclose all the information needed to reproduce the main experimental results of the paper to the extent that it affects the main claims and/or conclusions of the paper (regardless of whether the code and data are provided or not)?
    \item[] Answer: \answerYes{} 
    \item[] Justification: We provide detailed descriptions of the experimental setup, model architecture, hyperparameters, training procedure, and evaluation protocols.

\item {\bf Open access to data and code}
    \item[] Question: Does the paper provide open access to the data and code, with sufficient instructions to faithfully reproduce the main experimental results, as described in supplemental material?
    \item[] Answer: \answerNA{} 
    \item[] Justification: The paper reuses existing benchmark datasets and builds on top of previous open-source models.

\item {\bf Experimental setting/details}
    \item[] Question: Does the paper specify all the training and test details (e.g., data splits, hyperparameters, how they were chosen, type of optimizer, etc.) necessary to understand the results?
    \item[] Answer: \answerYes{} 
    \item[] Justification: We specify relevant training and testing details, including data splits, hyperparameters, and optimizer choices. These details align with those used in the reused code repositories for the baseline methods.

\item {\bf Experiment statistical significance}
    \item[] Question: Does the paper report error bars suitably and correctly defined or other appropriate information about the statistical significance of the experiments?
    \item[] Answer: \answerNo{} 
    \item[] Justification: Not all experiments include error bars or statistical significance measures due to the high computational cost of training some models, limiting repeated runs.

\item {\bf Experiments compute resources}
    \item[] Question: For each experiment, does the paper provide sufficient information on the computer resources (type of compute workers, memory, time of execution) needed to reproduce the experiments?
    \item[] Answer: \answerYes{} 
    \item[] Justification: The paper details the compute resources used of all experiments in~\cref{app:exp_details}.
    
\item {\bf Code of ethics}
    \item[] Question: Does the research conducted in the paper conform, in every respect, with the NeurIPS Code of Ethics \url{https://neurips.cc/public/EthicsGuidelines}?
    \item[] Answer: \answerYes{} 
    \item[] Justification: The research fully complies with the NeurIPS Code of Ethics in all aspects.

\item {\bf Broader impacts}
    \item[] Question: Does the paper discuss both potential positive societal impacts and negative societal impacts of the work performed?
    \item[] Answer: \answerNA{} 
    \item[] Justification: This work is foundational research without direct societal impact or applications that raise ethical concerns.
    
\item {\bf Safeguards}
    \item[] Question: Does the paper describe safeguards that have been put in place for responsible release of data or models that have a high risk for misuse (e.g., pretrained language models, image generators, or scraped datasets)?
    \item[] Answer: \answerNA{} 
    \item[] Justification: This paper does not release data or models with high risk for misuse.

\item {\bf Licenses for existing assets}
    \item[] Question: Are the creators or original owners of assets (e.g., code, data, models), used in the paper, properly credited and are the license and terms of use explicitly mentioned and properly respected?
    \item[] Answer: \answerYes{} 
    \item[] Justification: This work reuses existing assets with MIT license.

\item {\bf New assets}
    \item[] Question: Are new assets introduced in the paper well documented and is the documentation provided alongside the assets?
    \item[] Answer: \answerNA{} 
    \item[] Justification: This paper does not introduce or release any new assets.

\item {\bf Crowdsourcing and research with human subjects}
    \item[] Question: For crowdsourcing experiments and research with human subjects, does the paper include the full text of instructions given to participants and screenshots, if applicable, as well as details about compensation (if any)? 
    \item[] Answer: \answerNA{} 
    \item[] Justification: The paper does not involve crowdsourcing or any research with human subjects.

\item {\bf Institutional review board (IRB) approvals or equivalent for research with human subjects}
    \item[] Question: Does the paper describe potential risks incurred by study participants, whether such risks were disclosed to the subjects, and whether Institutional Review Board (IRB) approvals (or an equivalent approval/review based on the requirements of your country or institution) were obtained?
    \item[] Answer: \answerNA{} 
    \item[] Justification: This research does not involve human subjects or crowdsourcing.

\item {\bf Declaration of LLM usage}
    \item[] Question: Does the paper describe the usage of LLMs if it is an important, original, or non-standard component of the core methods in this research? Note that if the LLM is used only for writing, editing, or formatting purposes and does not impact the core methodology, scientific rigorousness, or originality of the research, declaration is not required.
    \item[] Answer: \answerNA{} 
    \item[] Justification: LLMs were not used in any core or non-standard part of the method. Their use, if any, was limited to minor writing edits and did not impact the research.

\end{enumerate}

\end{document}

%% file: DELTA_sections/introduction.tex
\section{Introduction}
\label{sec:intro}

Diffusion models have emerged as powerful methods for modeling complex distributions~\citepcolor{ho2020denoising, song2020denoising, song2020score, karras2022elucidating}, with applications in domains such as molecular and protein generation~\citepcolor{surveygraphdiff, digress, protein_diff}. Many physical systems, such as molecules or crystals, exhibit inherent symmetries. For example, a molecule's physical properties remain unchanged under rotations in 3D space~\citepcolor{hoogeboom2022equivariant, jing2022torsional}. Modeling such data requires learning distributions that are invariant under the action of a group~\(G\). This setting is naturally captured by the notion of \(\Gi\) distribution \(q(x_0)\) that are invariant under transformations from \(G\)~\citepcolor{Theoequi, equiflow}. 

Two main strategies have emerged for learning \(\Gi\) distributions: (1) designing equivariant network architectures, and (2) using data augmentation to approximate equivariance. Equivariant architectures, such as equivariant denoisers, ensure symmetry by construction~\citepcolor{hoogeboom2022equivariant, klein2024equivariant, igashov2024equivariant}, but are often less efficient due to increased architectural complexity and can pose optimization challenges~\citepcolor{doesequimatteratscale, nonuniversalityequi}. In contrast, data augmentation is a flexible and widely used alternative that approximates equivariant training by sampling transformed versions of the input. While this approach scales easily and has become increasingly popular in the community, especially for large-scale models~\citepcolor{alphafold3,proteina}, its effectiveness in capturing symmetries may vary depending on the group and application domain. In some settings, such as molecular dynamics and structural biology, explicit equivariance remains beneficial \citepcolor{protein_diff, batatia2022mace, zaverkinhigher}. In this work, we propose a framework that provably improves \emph{both} approaches: we introduce a novel form of implicit data augmentation by computing the denoising target as a weighted average over group orbits, which reduces variance and improves the training of equivariant denoisers.

We revisit data augmentation from a principled perspective and interpret it as a Monte Carlo estimator of the gradient of a symmetrized loss. This loss is defined over a fully symmetrized dataset, which yields an empirical distribution invariant under \(G\)~\citepcolor{Theoequi}. While traditional data augmentation uses one or few samples from this augmented dataset, we instead apply Rao–Blackwellization to derive a new estimator with provably lower variance. Specifically, we decompose the gradient as an outer expectation over data and an inner conditional expectation over group actions. Replacing the noisy sample-based target with its conditional expectation yields a lower-variance gradient estimator, while preserving equivariance.

To translate our theoretical insights into a practical method, we develop an efficient implementation of the proposed gradient estimator. This approach integrates the symmetrized loss and variance reduction into a modified training objective, without increasing the computational cost. Our implementation requires only a single forward and backward pass per sample, and it is compatible with both equivariant and non-equivariant architectures. We refer to this practical method as \textit{Orbit Diffusion}. By implicitly applying Rao–Blackwellization through a tailored loss formulation and sampling scheme, Orbit Diffusion enables stable optimization and improved generalization across a wide range of symmetry groups and tasks.

We provide theoretical guarantees that our symmetrized loss admits equivariant minimizers and that our gradient estimator has strictly smaller variance compared to existing methods. Empirically, we demonstrate strong performance across multiple domains. Our method achieves state-of-the-art results on GEOM-QM9 for molecular conformation generation, enhances crystal structure prediction, and improves text-guided crystal generation on the Perov-5 and MP-20 benchmarks. Moreover, our approach is compatible with non-equivariant denoisers; in particular, it improves the designability of protein structures generated by \textsc{ProteiNA}~\citepcolor{proteina}.

%% file: DELTA_sections/background.tex
\vspace{-0.23cm}
\section{Background}

\input{DELTA_sections/Background_long/group_brief}

\input{DELTA_sections/Background_long/diff}

%% file: DELTA_sections/Background_long/group_brief.tex
\paragraph{Groups.}
A \emph{group} is a mathematical structure comprising a set \( G \) and a binary operation \( m : G \times G \to G \) that combines two elements of \( G \). 
A \emph{group action} $g \in G$ defines how the group \( G \) acts on a set \( \Omega \), such as a set of geometric objects.\footnote{When the group acts on a vector space \( V \), we do not distinguish between the abstract group element \( g \in G \) and its linear representation \( \rho(g) : G \to \mathrm{GL}(V) \). For simplicity, we write \( g \circ x \)  to denote the action \( \rho(g)(x) \).} 
We restrict our attention to \emph{locally compact isometry groups}. Locally compact isometry groups encompass a broad class of transformation groups that preserve distances and possess a well-behaved topological structure. Examples include the permutation group \( \mathsf{S}_n \), the orthogonal group \( \mathsf{O}(d) \), and the special orthogonal group \( \mathsf{SO}(d) \).

\paragraph{Invariance and Equivariance.}
A function \( f : \Omega \to \mathbb{R} \) is said to be \emph{\( G \)-invariant} if for all \( g \in G \) and \( x \in \Omega \), it satisfies \( f(g \circ x) = f(x) \). This means the function value does not change under the action of any group element.
A function \( f : \Omega \to \Omega \) is said to be \emph{\( G \)-equivariant} if for all \( x \in \Omega \), \( f \) commutes with any group action \( g \in G \): \( f(g \circ x) = g \circ f(x) \).

\paragraph{Invariant and Equivariant Distributions.} 
A probability distribution \( p(x) \) defined on a set \( \Omega \) is said to be \(\Gi\) under the action of a group \( G \) if the probability of any measurable subset \( A \subseteq \Omega \) remains unchanged under the transformation induced by any group action \( g \in G : p(g \circ x \in A) = p(x \in A). \) A conditional distribution \( p(y \mid x) \), where \( x \in \Omega \) and \( y \in \Omega \), is said to be \(\Ge\) if for all \( g \in G \), the following condition holds: \(
p(g \circ y \mid g \circ x) = p(y \mid x). \). 

\paragraph{Group Symmetrization.}  
Let \( S_G \) be the symmetrization operator under group \( G \), transforming any distribution \( p(x) \) into a \(\Gi\)-invariant distribution, denoted as the \(\Gs\) distribution:
\begin{equation}
    \label{eqn:group_symetrization}
    S_G[p](x) := \int_G p(g \circ x) \, \mathrm{d}\mu_G(g), \quad \text{ where } \mu_G \text{ is the Haar measure on } G.
\end{equation}

%% file: DELTA_sections/Background_long/diff.tex
\vspace{-0.3cm}
\paragraph{Diffusion Models and Equivariant Diffusion Models.}
Diffusion models~\citepcolor{ho2020denoising} are a class of generative models that construct complex data distributions by iteratively transforming simple noise distributions through a learned denoising process. 
Formally, given data \( x_0 \sim q(x_0) \), the forward process generates a sequence \( x_t \) over time \( t \in [0, T] \) using a stochastic differential equation (SDE)~\citepcolor{song2020score} or a discrete Markov chain~\citepcolor{ho2020denoising}, such as:
\[
x_t = {\alpha_t} x_0 + \sigma_t \epsilon, \quad \epsilon \sim \mathcal{N}(0, I),
\]
where \( \alpha_t \) and \( \sigma_t \) are a time-dependent scaling factor and a noise factor that determines the level of noise added at each step, respectively. The noise should be increasingly added to the sample so that at time $t = T$, $q(x_T) \approx \mcN(0, I)$.

The reverse process, parameterized by a neural network \( \phi_\theta(x_t, t) \), approximates a clean sample \( x_0 \) given its noisy version \( x_t \). The training objective typically involves minimizing a reweighted form of the denoising loss~\citepcolor{ho2020denoising, song2020score, karras2022elucidating}:
\begin{equation}
    \mcL = \mathbb{E}_{t \sim \mathcal{U}(0, T), (x_0, x_t) \sim q(x_0, x_t)} \left[ \omega(t) \| \phi_\theta (x_t, t) - x_0 \|^2 \right],
    \label{eqn:diff_loss}
\end{equation}
where \(\omega(t)\) is a time-dependent loss weight.  For notational simplicity, we omit this term throughout the remainder of the paper. To sample from the diffusion model, we begin with a noise vector \( x_T \sim \mathcal{N}(0, I) \) and iteratively apply the learned reverse process to transform it into a data sample \( x_0 \) using the trained model \( \phi_\theta \). The reverse process can involve solving an ODE or SDE numerically~\citepcolor{song2020score, karras2022elucidating, lu2022dpm, tong2024learning}.

Equivariant diffusion models extend standard diffusion by enforcing equivariance of the neural network denoiser. Specifically, the denoiser \(\phi_\theta\) is said to be \(\Ge\) if it satisfies \(\phi_\theta(g \circ x_t, t) = g \circ \phi_\theta(x_t, t)\) for all \(g \in G\). 



%% file: AI4Science/method_reformulate.tex
\section{Method}
\label{sec:method}

\begin{figure}[ht]
\centering
\begin{subfigure}[T]{.5055\textwidth}
\includegraphics[width=1.0\linewidth]{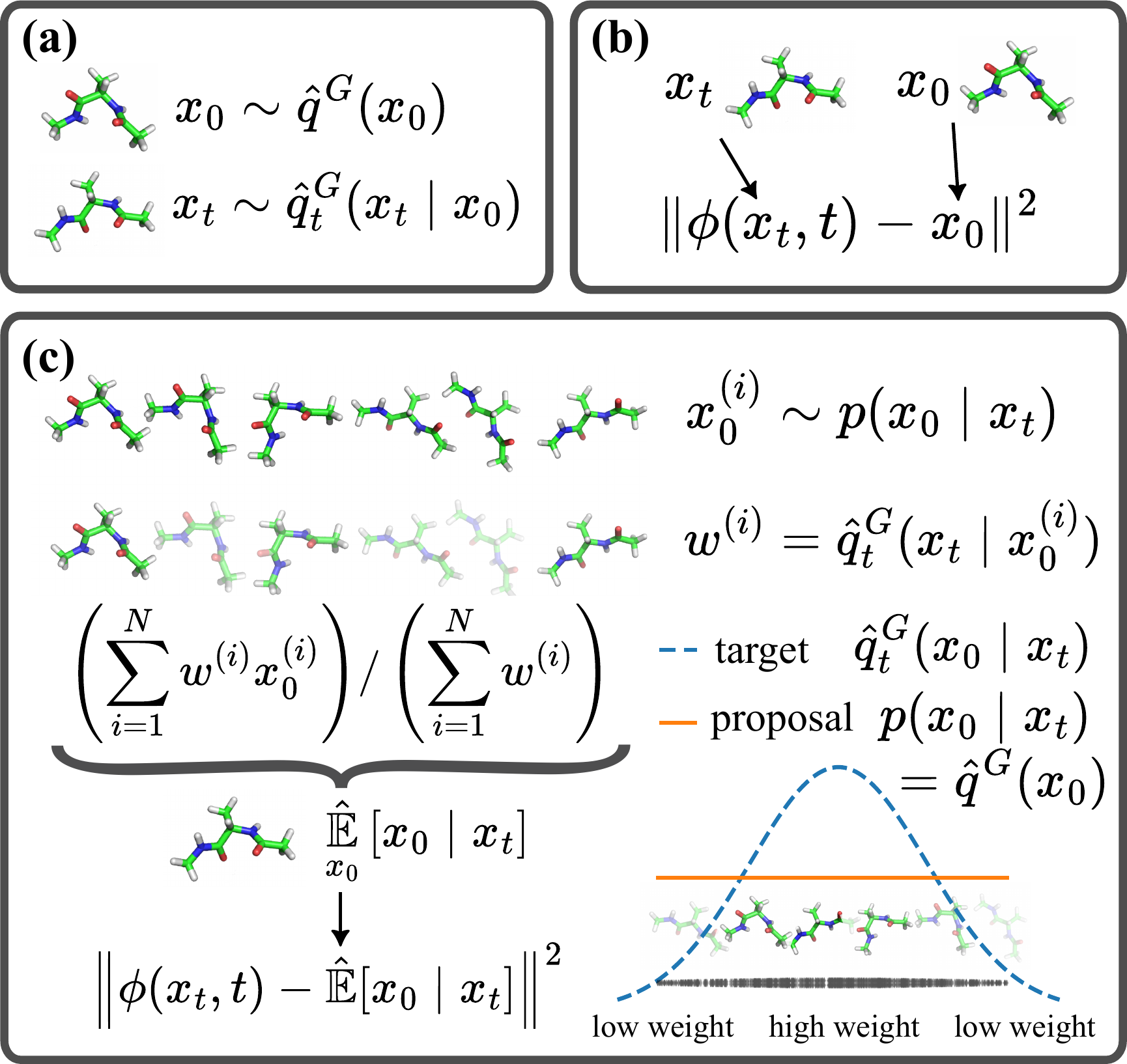}
\label{fig:illustration}
\end{subfigure}%
\hfill
\begin{subfigure}[T]{.474\textwidth}
\small
\begin{tcolorbox}[title=Algorithm 1: Orbit Diffusion with RB.,boxsep=1pt,left=2pt,right=2pt,top=2pt,bottom=2pt,colback=white]
\label{algo:rb-estimator}
\begin{algorithmic}[1]
\State Sample a data point $x_0 \sim \hat{q}^G(x_0)$
\State Sample a noise level $t$
\State Generate a noisy sample $x_t \sim \hat{q}_t^G(x_t \mid x_0)$
\For{$i = 1$ to $N$}
    \State Sample a group element $g^{(i)} \sim \nu_t(g)$
    \State Compute orbit sample $x_0^{(i)} = g^{(i)} \circ x_0$
    \State Compute $w^{(i)} = \hat{q}_t^G(x_t \mid x_0^{(i)})/\nu_t(g^{(i)})$
\EndFor
\State Approximate $\mathbb{E}[x_0 \mid x_t]$ with SNIS:
\[  \hat{\mathbb{E}}[x_0 \mid x_t] = \left( \sum_{i=1}^N w^{(i)} x_0^{(i)} \right) / \left(\sum_{i=1}^N w^{(i)}\right)
\]
\State Backpropagate gradients of the loss: 
\[ 
\left\| \phi(x_t, t) - \hat{\mathbb{E}}[x_0 \mid x_t] \right\|^2
\]
\end{algorithmic}
\end{tcolorbox}
\label{fig:algorithm}
\end{subfigure}

\caption{Gradient estimation strategies for training (approximately) equivariant diffusion models:
(a) Sampling from the symmetrized joint distribution to obtain \( x_0 \) and \( x_t \).
(b) The standard data augmentation approach, which directly uses these samples for training.
(c) The proposed method, leveraging self-normalizing importance sampling (SNIS) to estimate the inner conditional expectation.
Both (b) and (c) require a single neural function evaluation per gradient step, but (c) has lower variance than (b). The pseudo-code for the Rao-Blackwell estimator with SNIS is shown on the right.
}

\label{fig:illustration-and-algorithm}
\end{figure}

Let \( G \) be a symmetry group, such as the group of Euclidean rotations.  
Our goal is to learn a \( G \)-invariant data-generating distribution \( q(x_0) \).  
However, the observed distribution \(\hat{q}\) (from which we obtain training samples) is generally not \(G\)-invariant due to dataset biases.  
For example, in molecular datasets, each molecule may be stored in a canonical but arbitrary orientation, even though physically all rotated versions are equally probable under \(q\).  
As a result, training directly on \(\hat{q}\) would lead to a model that may not respect the underlying symmetry.  

This issue is widely recognized in the literature, and two standard solutions are:
\begin{enumerate}
    \item \textbf{Equivariant model:} Train with the diffusion loss in \cref{eqn:diff_loss} using a \(G\)-equivariant network \( \phi(x_t, t) \).
    \item \textbf{Data augmentation:} Train a non-equivariant \( \phi \) with \cref{eqn:diff_loss}, augmenting data with random group actions to encourage approximate \( G \)-invariance.
\end{enumerate}

The second approach has been widely adopted—several high-profile non-equivariant models achieve strong results through augmentation~\citepcolor{alphafold3,proteina}.  
However, empirical evidence shows that augmentation offers \emph{no benefit} for already equivariant models, a result we formally prove in \cref{sub_app:equivalance}.

\subsection{From Symmetrized Loss to High-Variance Gradient Estimators}

To unify these approaches, consider the \emph{symmetrized data distribution} from \cref{eqn:group_symetrization},
and the forward noising kernel \(\hat{q}_t^G(x_t \mid x_0)\) (e.g., a Gaussian in denoising diffusion) with marginal \(\hat{q}_t^G(x_t)\).  
The \emph{symmetrized diffusion loss} at time \(t\) is
\begin{align}
    \mathcal{L}^G_t(\phi) 
    &= \mathbb{E}_{x_0 \sim \hat{q}^G} \, \mathbb{E}_{x_t \sim \hat{q}_t^G(\cdot \mid x_0)} 
    \big[ \| \phi(x_t, t) - x_0 \|^2 \big].
    \label{eqn:sym_loss}
\end{align}
Both the equivariant model and the data augmentation approach can be viewed as implicitly minimizing \(\mathcal{L}^G_t(\phi)\) \citepcolor{Theoequi}.  
The \emph{true} gradient of this loss is
\begin{align}
    \nabla_\phi \mathcal{L}^G_t
    &= \mathbb{E}_{x_t \sim \hat{q}_t^G} \, 
       \mathbb{E}_{x_0 \sim \hat{q}_t^G(\cdot \mid x_t)} 
       \big[ 2\big( \phi(x_t, t) - x_0 \big) \big],
    \label{eqn:sym_gradient}
\end{align}
where the expectations are taken over the full symmetrized joint distribution.

In practice, we do not have access to the exact expectations in \cref{eqn:sym_gradient}. Instead, we construct a Monte Carlo \emph{gradient estimator} by sampling \(x_0 \sim \hat{q}\) (or \(\hat{q}^G\) in the augmented/equivariant case), then sampling \(x_t \sim \hat{q}^G_t(\cdot \mid x_0)\), and using the single-sample estimate
\(
\widehat{\nabla_\phi \mathcal{L}^G_t} = 2 \big( \phi(x_t, t) - x_0 \big).
\)
This estimator is \emph{unbiased}, but, as in other diffusion training setups, it can exhibit \emph{high variance}~\citepcolor{kingma2021variational, xu2023stable, nichol2021improved}.  
Increasing the batch size reduces variance but requires more forward passes of the neural network \( \phi \), raising computational cost.

In this work, we introduce a class of Rao--Blackwellized gradient estimators that \emph{provably reduce variance} while remaining unbiased, and can be applied to both the equivariant and augmentation-based training strategies.

\subsection{Rao--Blackwellized Gradient Estimator}
Our \emph{key observation} is that \(\phi(x_t, t)\) does not depend on \(x_0\). This allows us to move it outside the inner expectation in \cref{eqn:sym_gradient}, yielding
\begin{align}
    \nabla_\phi \mathcal{L}^G_t
    &= \mathbb{E}_{x_t \sim \hat{q}_t^G} 
    \Big[ 2\big( \phi(x_t, t) - \underbrace{\mathbb{E}_{x_0 \sim \hat{q}^G_t(x_0 \mid x_t)}[x_0]}_{\mathbb{E}[x_0 \mid x_t]} \big) \Big].
    \label{eqn:new_sym_grad}
\end{align}
Replacing \(x_0\) with its conditional mean \(\mathbb{E}[x_0 \mid x_t]\) yields a \emph{Rao--Blackwellized (RB) gradient estimator}, which remains unbiased and has variance no greater than the original—strictly less unless \(x_0 \mid x_t\) is deterministic, a situation that rarely occurs in generative modeling where \(x_0\) is typically stochastic given \(x_t\). This improvement requires no additional neural network evaluations, only a more accurate target. To avoid custom backward passes, we can minimize:
\begin{align}
\mathcal{L}_t^{\mathrm{RB}}(\phi) 
&= \mathbb{E}_{x_t \sim \hat{q}_t^G}  
\left[ \left\| \phi(x_t, t) - \mathbb{E}[x_0 \mid x_t] \right\|^2 \right] 
\label{eqn:RB_sym_loss}
\end{align}

\begin{tcolorbox}[title=Variance reduction guarantee]
\begin{theorem}
Let \(\widehat{\nabla}_\phi\) be the Monte Carlo gradient from \cref{eqn:sym_gradient} and \(\widehat{\nabla}_\phi^{(RB)}\) be from \cref{eqn:new_sym_grad}.  
If \(\mathbb{E}[x_0 \mid x_t]\) can be computed exactly, then
\[
\mathrm{Var}\big(\widehat{\nabla}_\phi^{(RB)}\big) \le \mathrm{Var}\big(\widehat{\nabla}_\phi\big),
\]
with strict inequality unless \(x_0 \mid x_t\) is a Dirac delta.
\label{the:theo_1}
\end{theorem}
\end{tcolorbox}

The challenge now is estimating the loss target \(\mathbb{E}[x_0 \mid x_t]\) accurately and efficiently.  
We address this next using self-normalized importance sampling (SNIS).

\subsection{Estimating the Conditional Expectation} 
A central challenge in computing the gradient estimator is evaluating the conditional expectation \( \mathbb{E}_{x_0 \sim \hat{q}^G_t(x_0 \mid x_t)}[x_0] \), which is generally intractable:
\begin{align}
    \mathbb{E}_{x_0 \sim \hat{q}^G_t(x_0 \mid x_t)}[x_0] = \int_{\Omega} x_0 \, \hat{q}_t^G(x_0 \mid x_t) \, \mathrm{d}x_0.
\end{align}
This expectation can be approximated by drawing independent samples \(x_0^{(1)}, \dots, x_0^{(N)} \sim \hat{q}^G_t(x_0 \mid x_t)\) and computing the sample mean. Unfortunately, we cannot sample efficiently and directly from \( \hat{q}^G_t(x_0 \mid x_t) \). Using Bayes' rule:
\begin{align}
    \hat{q}_t^G(x_0 \mid x_t) \propto \hat{q}_t^G(x_t \mid x_0) \hat{q}^G(x_0),
\end{align}
where \( \hat{q}_t^G(x_t \mid x_0) \) is available in closed form, but \( \hat{q}^G(x_0) \) is intractable due to integration over the group orbit of \(x_0\).

To address this, we use self-normalized importance sampling (SNIS) with a proposal distribution \( p(x_0 \mid x_t) \) that shares the same intractable orbit-integral structure as \( \hat{q}^G(x_0) \), allowing cancellation of the problematic terms in the importance weights. The conditional expectation is approximated as:
\begin{align}
    \mathbb{E}_{x_0 \sim \hat{q}^G_t(x_0 \mid x_t)}[x_0] 
    \approx \frac{ \sum_{i=1}^N x_0^{(i)} \cdot w^{(i)} }{ \sum_{i=1}^N w^{(i)} }, \quad
    w^{(i)} = \hat{q}_t^G(x_t \mid x_0^{(i)}) \cdot \frac{\hat{q}^G(x_0^{(i)})}{p(x_0^{(i)} \mid x_t)}.
\end{align}
The design of the proposal \( p(x_0 \mid x_t) \) aims to ensure that the quotient  \( \hat{q}^G(x_0) / p(x_0 \mid x_t) \) becomes tractable. This can be achieved by first sampling from the original dataset \( D \) using some user-defined \( \bar{p}(x_0 \mid x_t) \), where \( \bar{p} \) is designed to have non-zero probability for all elements in the dataset. Once a sample \( x_0 \in D \) is drawn, we sample a group element \( g \) from the group \( G \) uniformly at random, and apply this group action to the sample. This results in a new sample \( x_0^{(i)} = g \circ x_0 \). This proposal distribution inherits the same orbit-integral structure as \( \hat{q}^G(x_0) \), causing the intractable terms in the ratio  \( \hat{q}^G(x^{(i)}_0) / p(x^{(i)}_0 \mid x_t) \) to cancel. Specifically, with $\delta$ the Dirac delta function:
\begin{align}
    \frac{\hat{q}^G(x_0^{(i)})}{p(x_0^{(i)} \mid x_t)} = \frac{\hat{q}^G(g \circ x_0)}{p(g \circ x_0 \mid x_t)} = \frac{\hat{q}(x_0) \int_G \delta(g \circ x_0 - g' \circ x_0) \, \mathrm{d}\mu_G(g')}{\bar{p}(x_0 \mid x_t) \int_G \delta(g \circ x_0 - g' \circ x_0) \, \mathrm{d}\mu_G(g')} = \frac{\hat{q}(x_0)}{\bar{p}(x_0 \mid x_t)}.
\end{align}

Beside, since \(\hat{q}(x_0) = 1/|D|\) for any \(x_0 \in D\), this term can be omitted from the importance weight. Thus, the final importance weight simplifies to \( w^{(i)} =   {\hat{q}_t^G(x_t \mid x_0^{(i)})}/{\bar{p}(x_0 \mid x_t)}\). Importantly, all components of the importance weights are tractable: \( \hat{q}_t^G(x_t \mid x_0^{(i)}) \) is the forward diffusion process; \( \hat{q}(x_0) \) corresponds to the empirical data distribution; and \( \bar{p}(x_0 \mid x_t) \) is user-defined and tractable. Moreover, SNIS estimators based on these importance weights are always consistent.

An important instance results from setting \( \bar{p}(x_0 \mid x_t) = \hat{q}(x_0) \) where the proposal recovers the exact symmetrized distribution: \( p(x_0 \mid x_t) = \hat{q}^G(x_0) \) and the importance weight simplifies to \( w^{(i)} = \hat{q}_t^G(x_t \mid x_0^{(i)}) \).

\subsection{Practical Implementation—Orbit Diffusion (\texttt{OrbDiff})}  
We present \emph{Orbit Diffusion} (\texttt{OrbDiff}) as a practical variant of our estimator.  
Although using $\hat{q}^G(x_0)$ as the proposal is theoretically valid, it is inefficient and cumbersome in practice. For small $t$, the conditional $\hat{q}_t^G(x_t \mid x_0)$ is sharply concentrated around the $x_0$ that generated $x_t$, so uniformly sampled $x_0$ rarely yield useful gradients. Furthermore, sampling from the full support generally requires drawing points outside the current minibatch, adding non-trivial implementation complexity.

To improve efficiency, we fix $x_0$ to the example that produced $x_t$ and sample candidates only from its \emph{orbit} $\mathcal{O}_{x_0} = \{ g \circ x_0 \mid g \in G \}$.  
This biases the proposal toward points with high likelihood under $\hat{q}_t^G(x_t \mid x_0)$---for instance, small rotations in $\mathsf{SO}(3)$-equivariant settings or local permutations in discrete symmetry groups. Since such points dominate the conditional distribution at small noise levels, orbit sampling greatly improves sample efficiency by prioritizing candidates with non-trivial importance weights. At high noise levels, contributions from outside the orbit may increase, but their weights are typically small, and expanding the proposal has shown little benefit.

Formally, \texttt{OrbDiff} replaces the intractable conditional expectation $\mathbb{E}[x_0 \mid x_t]$ in \cref{eqn:RB_sym_loss} with the \emph{orbit-weighted target}  
\begin{align}
\phi^*(x_0, x_t, t) 
= \frac{1}{Z(x_t, x_0)} \int_G (g \circ x_0)\, \hat{q}^G_t(x_t \mid g \circ x_0)\, \mathrm{d}\mu_G(g),
\label{eqn:weight_orbit_integral}
\end{align}
where $Z(x_t, x_0) = \int_G \hat{q}_t^G(x_t \mid g \circ x_0)\, \mathrm{d}\mu_G(g)$ is the normalization constant.  
This yields the \texttt{OrbDiff} loss:
\begin{align}
\mathcal{L}_t^{\texttt{OrbDiff}}(\phi) 
= \mathbb{E}_{x_0 \sim \hat{q}^G(x_0)} 
\mathbb{E}_{x_t \sim \hat{q}_t^G(\cdot \mid x_0)} 
\left[ \left\| \phi(x_t, t) - \phi^*(x_0, x_t, t) \right\|^2 \right].
\label{eqn:final_sym_loss}
\end{align}

\cref{fig:illustration-and-algorithm} provides an illustration and pseudo-code of the practical implementation. When $\phi$ is $G$-equivariant (see \cref{sub_app:equivalance} and \citetcolor{Theoequi}), the loss in \cref{eqn:final_sym_loss} and its gradient are unchanged if $x_0$ is drawn from the empirical distribution $\hat{q}(x_0)$ rather than $\hat{q}^G(x_0)$.

Even though $\phi^*(x_0, x_t, t) \neq \mathbb{E}[x_0 \mid x_t]$, for an equivariant forward process the gradient of \cref{eqn:final_sym_loss} matches that of \cref{eqn:new_sym_grad}, ensuring that \texttt{OrbDiff} yields an unbiased gradient estimate. The orbit-weighted target is also equivariant, providing a training signal aligned with the model’s inductive bias. We formally prove both properties in \cref{sub_app:orbit_expectation}.

\begin{tcolorbox}[title=Unbiased gradient and equivariance of \texttt{OrbDiff}]
\begin{theorem}
\label{theo_2}
Let \( G \) be a locally compact isometry group acting on data space \( \Omega \), and suppose the forward kernels \( \hat{q}_t^G(x_t \mid x_0) \) are \( G \)-invariant: \( \hat{q}_t^G(g \circ x_t \mid g \circ x_0) = \hat{q}_t^G(x_t \mid x_0) \) for all \( g \in G \). Then:

\begin{enumerate}
    \item The \texttt{OrbDiff} target
\(
\phi^*(x_0, x_t, t)
\)
satisfies: \( \phi^*(x_0, h \circ x_t, t) = h \circ \phi^*(x_0, x_t, t) \) for all \( h \in G \).

    \item The gradient of the \texttt{OrbDiff} loss \eqref{eqn:final_sym_loss} equals that of the ideal loss \eqref{eqn:new_sym_grad}, i.e., \texttt{OrbDiff} provides an unbiased gradient estimator.
\end{enumerate}
\end{theorem}
\end{tcolorbox}

We also explore non-uniform sampling of group elements \(g\) for approximating the conditional expectation. For small noise, we sample near the identity action, expanding the neighborhood as noise increases. These distributions, \(\nu_t(g)\), depend on the noise schedule. While not all groups support closed-form expressions for the density of individual group elements, they exist for the translation group (sampled from a Gaussian) or \(\mathsf{SO}(3)\) (sampled from the von Mises-Fisher distribution).

Next, we divide by \(\nu_t(g)\) to account for the group sampling distribution, resulting in the importance weights
\(w^{(i)} = {\hat{q}_t^G(x_t \mid x_0^{(i)})}/{\nu_t(g)}\). We also ensure that the identity group element is included in the sampled set. This strategy is effective in practice and provides computational advantages.

%% file: DELTA_sections/experiment.tex
\section{Experimental Results}
\label{sec:exp}

Our experiments evaluate the generality and robustness of Orbit Diffusion across diverse generative tasks. We begin with a controlled synthetic setup using a standard diffusion model (\cref{exp:synthetic}) and consider various isometry groups, including reflections, rotations, translations, and graph automorphisms. We then extend our method to Flow Matching (\cref{exp:molecule}) and to diffusion models with non-standard forward processes (\cref{exp:crystal}). Finally, we apply Orbit Diffusion to a non-equivariant denoiser, demonstrating its effectiveness without architectural symmetry (\cref{exp:proteina}).

\vspace{-0.2cm}
\input{DELTA_sections/Experiments/toy}

\vspace{-0.2cm}
\input{DELTA_sections/Experiments/conformer}
\vspace{-0.2cm}
\input{DELTA_sections/Experiments/diffcsp}
\vspace{-0.2cm}
\input{DELTA_sections/Experiments/protein}
\vspace{-0.2cm}
\input{DELTA_sections/Experiments/analysis}

%% file: DELTA_sections/Experiments/toy.tex
\subsection{Experiments on Synthetic Data}
\label{exp:synthetic}

\input{DELTA_sections/Experiments/figure_1d}

We construct a dataset with a single 1D sample \( x_0 = 1 \), where the equivariant group is reflection: \( g \circ x = \pm x \). We train a denoiser (EquiNet) of the form \( D_\theta(x_t, t) = f_\theta(x_t, t) - f_\theta(-x_t, t) \), which is equivariant by design, with \( f_\theta \) being a simple 3-layer MLP.

We evaluate three training variants of EquiNet: default, trained without Orbit Diffusion or data augmentation; + [Aug], with data augmentation only; and + [\texttt{OrbDiff}], with Orbit Diffusion only. Each model is trained for 100k iterations. After training, we generate 100k samples and compute the root mean square deviation (RMSD) from the closest target in \(\{-1, 1\}\), and the Wasserstein-2 (W2) distance to the target distribution. We report mean and standard deviation for each model in~\cref{tab:1d_toy}.

\Cref{fig:training_loss_nolog} shows the training loss curves for all three EquiNet variants. EquiNet and its augmented version exhibit similar fluctuations and magnitudes, indicating that data augmentation does not reduce variance. This is consistent with our theoretical result (\cref{sub_app:equivalance}): for an equivariant denoiser, augmented and non-augmented losses are equivalent. In contrast, the \texttt{OrbDiff} variant shows a smoother and lower loss curve, confirming that Rao-Blackwellization reduces gradient variance and stabilizes training. As shown in \cref{tab:1d_toy}, \texttt{OrbDiff} achieves roughly 25× lower RMSD and 200× lower W2 distance, outperforming both EquiNet variants.

%% file: DELTA_sections/Experiments/figure_1d.tex
\begin{figure}[ht]
    \centering
    \begin{minipage}{0.33\textwidth}
        \vspace{-0.39cm}
        \includegraphics[]{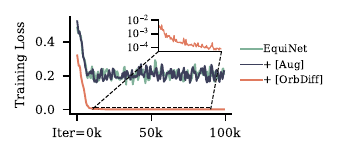}
        \vspace{-0.6cm}
        \caption{Learning curves.}
        \label{fig:training_loss_nolog}
    \end{minipage}
    \hfill
    \begin{minipage}{0.64\textwidth}
        \centering
            \setlength{\tabcolsep}{1.4pt}
    \renewcommand{\arraystretch}{0.9}
        \resizebox{0.75\textwidth}{!}{ 
        \begin{tabular}{lcc}
            \toprule
            Variant & RMSD ($\times 10^{-5}$)~(\( \downarrow \)) & W2 ($\times 10^{-3}$)~(\( \downarrow \)) \\
            \midrule
            EquiNet & 9.05 $\pm$ 9.47 & 1.150 $\pm$ 1.581 \\
            + [Aug] & 9.50 $\pm$ 9.63 & 0.853 $\pm$ 1.156 \\
            \rowcolor[HTML]{EAEAEA}
            + [\texttt{OrbDiff}] & \textbf{0.37 $\pm$ 0.09} & \textbf{0.004 $\pm$ 0.001} \\
            \bottomrule
        \end{tabular}
        }
        \captionof{table}{Synthetic experiment results: RMSD to the closest target in \(\{-1, 1\}\) and W2 distance to ground-truth distribution.}
        \label{tab:1d_toy}
    \end{minipage}
    \vspace{-0.5em}
\end{figure}

%% file: DELTA_sections/Experiments/conformer.tex
\subsection{Molecular Conformer Generation}
\label{exp:molecule}

\paragraph{Molecular Conformer Generation.} 
Molecular Conformer Generation aims to generate plausible 3D structures from 2D molecular graphs, crucial for drug discovery and property prediction due to the role of 3D geometry~\citepcolor{Liu2022MolecularGPA, Axelrod2020MolecularMLA}.

We evaluate on the \textsc{GEOM-QM9} dataset~\citepcolor{axelrod2022geom}, respecting two key symmetries: invariance under global 3D rotations and equivariance to graph automorphisms, which permute atom indices without altering molecular identity. We compare against strong baselines, including \textsc{GeoMol}~\citepcolor{geomol}, Torsional Diffusion~\citepcolor{jing2022torsional}, \textsc{MCF}~\citepcolor{Swallow}, and \textsc{ETFlow}~\citepcolor{hassan2024flow}. Our method, Orbit Diffusion, is integrated into \textsc{ETFlow}, a strong equivariant flow matching model that employs a harmonic prior for bonded atom proximity. We finetune ETFLow with \texttt{OrbDiff} using their public checkpoint.




During training, we apply symmetry-aware sampling by uniformly sampling 50 automorphisms and 200 \(\mathsf{SO}(3)\) rotations per molecule, including the identity. These are applied to both 2D graphs and 3D conformers. All other settings follow \textsc{ETFlow}; see~\cref{sup_app:mcg} for details.

We benchmark against three versions of \textsc{ETFlow}: the results reported in the original paper, their released checkpoint, and our own reproduced results using the provided code and configuration \footnote{\url{https://github.com/shenoynikhil/ETFlow}}. Despite extensive effort, we were unable to match their reported performance, so we report all results under the same evaluation protocol. 
\vspace{-0.85cm}

\input{DELTA_sections/Experiments/table_and_viz_qm9}
\vspace{-0.45cm}
Table~\ref{tab:qm9_} shows our method consistently improves both precision and diversity. \texttt{OrbDiff} achieves the best recall scores, including a 4\% improvement in mean Cov@0.1, and the lowest Recall AMR (mean and median). It also maintains competitive precision at 0.1~\AA. While \textsc{MCF} performs better at 0.5~\AA\ precision, \texttt{OrbDiff} achieves the lowest AMR overall. Further experimental details and comparisons with more baselines are in~\cref{sup_app:mcg}.

%% file: DELTA_sections/Experiments/table_and_viz_qm9.tex
\begin{figure}[H] 
    \centering
    \begin{minipage}[b][][b]{0.68\linewidth}       
        \begin{table}[H]
            \caption{Molecular conformer generation performance on GEOM-QM9.  
            \footnotesize  
            \textsuperscript{*} Reported in the original paper.  
            \textsuperscript{†} Obtained using the published checkpoint.  
            \textsuperscript{‡} We train the public implementation from scratch.  
            }
            \label{tab:qm9_}
            \centering
            \setlength{\tabcolsep}{1.4pt}
            \renewcommand{\arraystretch}{0.9}
            \resizebox{\textwidth}{!}{
            \begin{tabular}{lcccccccccccc}
            \toprule
            \multirow{2}{*}{Models} & \multicolumn{6}{c}{Recall} & \multicolumn{6}{c}{Precision} \\
            \cmidrule(lr){2-7} \cmidrule(lr){8-13}
             & \multicolumn{2}{c}{Cov@0.1 (↑)} & \multicolumn{2}{c}{Cov@0.5 (↑)} & \multicolumn{2}{c}{AMR (↓)} 
             & \multicolumn{2}{c}{Cov@0.1 (↑)} & \multicolumn{2}{c}{Cov@0.5 (↑)} & \multicolumn{2}{c}{AMR (↓)} \\
            \cmidrule(lr){2-3} \cmidrule(lr){4-5} \cmidrule(lr){6-7} \cmidrule(lr){8-9} \cmidrule(lr){10-11} \cmidrule(lr){12-13}
             & Mean & Median & Mean & Median & Mean & Median & Mean & Median & Mean & Median & Mean & Median \\
            \midrule
            \textsc{GeoMol\textsuperscript{†} } & 28.4 & 0.0 & 91.1 & \textbf{100.0} & 0.224 & 0.194 & 20.7 & 0.0 & 85.8 & \textbf{100.0} & 0.271 & 0.243 \\
            Torsional Diff.\textsuperscript{†}  & 37.7 & 25.0 & 88.4 & \textbf{100.0} & 0.178 & 0.147 & 27.6 & 12.5 & 84.5 & \textbf{100.0} & 0.221 & 0.195 \\
            \textsc{MCF\textsuperscript{†} } & 81.9 & \textbf{100.0} & 94.9 & \textbf{100.0} & 0.103 & 0.049 & 78.6 & 93.8 & \textbf{93.9} & \textbf{100.0} & \textbf{0.113} & 0.055 \\
            \midrule
            \textsc{ETFlow\textsuperscript{*}} & - & - & 96.5 & \textbf{100.0} & 0.073 & 0.047 & - & - & 94.1 & \textbf{100.0} & 0.098 & 0.039 \\
            \textsc{ETFlow\textsuperscript{†}  } & 79.5 & \textbf{100.0} & 93.8 & \textbf{100.0} & 0.096 & 0.037 & 74.4 & 83.3 & 88.7 & \textbf{100.0} & 0.142 & 0.066 \\
            \textsc{ETFlow\textsuperscript{‡} } & 81.4 & \textbf{100.0} & 94.4 & \textbf{100.0} & 0.092 & 0.039 & 74.6 & 85.5 & 89.1 & \textbf{100.0} & 0.145 & 0.064 \\
        
            \rowcolor[HTML]{EAEAEA}
            + {[\texttt{OrbDiff}]} & \textbf{{85.4}} & \textbf{100.0} & \textbf{{96.3}} & \textbf{100.0} & \textbf{{0.074}} & \textbf{0.027} & \textbf{{80.2}} & \textbf{{93.9}} & {91.9} & \textbf{100.0} & \textbf{{0.113}} & \textbf{0.042} \\
            \bottomrule
            \end{tabular}}
        \end{table}
    \end{minipage}%
    \hfill
    \begin{minipage}[b][][b]{0.3\linewidth}
        \vspace{-0.5cm}
        \includegraphics[width=0.97\linewidth]{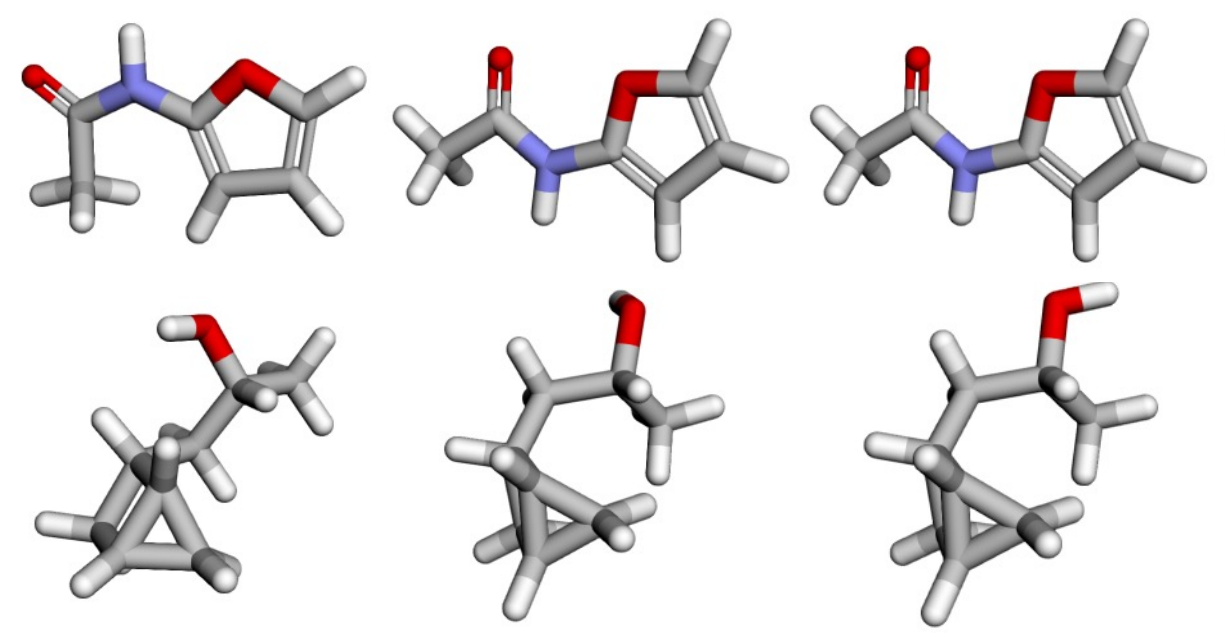}
        \caption{Molecular conformers generated by \textsc{ETFlow} (left), + [\texttt{OrbDiff}] (center), and ground-truth (right).}
        \label{fig:qm9_viz}
    \end{minipage}
\end{figure}

%% file: DELTA_sections/Experiments/diffcsp.tex
\subsection{Crystal Structure Prediction (CSP)}
\label{exp:crystal}
 
CSP involves recovering 3D atomic positions and lattice parameters from chemical composition. Due to periodicity, it suffices to predict the structure within a single unit cell, where coordinates lie in the fractional domain \([0,1)^{3 \times M}\). To handle this, DiffCSP uses a Wrapped Normal diffusion that respects periodic translation symmetry. We integrate Orbit Diffusion into DiffCSP, demonstrating that our approach extends beyond Gaussian diffusion models. We test two variants: {\texttt{OrbDiff\_U}}, which samples uniformly over the translation group, and {\texttt{OrbDiff\_WN}}, a time-dependent Wrapped Normal centered at zero, concentrating around \({x}_0\) at low noise and spreading out at high noise. Details of the \texttt{OrbDiff\_WN} proposal are in~\cref{sub_app:csp_wn}.

We evaluate our method on two CSP benchmarks: Perov-5~\citepcolor{castelli2012new, castelli2012computational} and MP-20~\citepcolor{jain2013commentary}. We use the three strongest baselines from the DiffCSP paper~\citepcolor{jiao2023crystal}: P-cG-SchNet~\citepcolor{diffcsp_bl1}, CDVAE~\citepcolor{diffcsp_bl2}, and DiffCSP, all with publicly available implementations. We evaluate performance using two standard metrics: Match Rate (the proportion of correctly matched structures in the test set) and RMSD (the average atomic deviation for matched samples, normalized by lattice volume). Full metric definitions and details are in Appendix~\ref{sub_app:CSP_metrics}.

We further consider a relevant task, introduced by TGDMat \citepcolor{das2025periodic}, where crystal structures are generated conditioned on additional text descriptions of the desired structures. In this task, two types of descriptions are considered: long and short, with the latter being easier to obtain than the former. We follow the same evaluation framework as for Non-text-guided CSP.

\input{DELTA_sections/Experiments/table_diffcsp}

From Tables ~\ref{tab:tgdmat} and ~\ref{tab:diffcsp}, one can see {\texttt{OrbDiff\_WN}} consistently enhances the performance in all cases, with a notable increase from 59.39\% to 65.57\% for TGDmat (S) on Perov-5 and from 61.91\% to 66.74\% for TGDMat (L) on MP-20. At the same time, {\texttt{OrbDiff\_U}} outperforms the baselines in 5 out of 6 cases, showing consistent benefits. Improvements are also observed consistently in RMSE.
\vspace{-0.1cm}

\begin{figure}[!ht]
    \centering
    \includegraphics[width=1.0\linewidth]{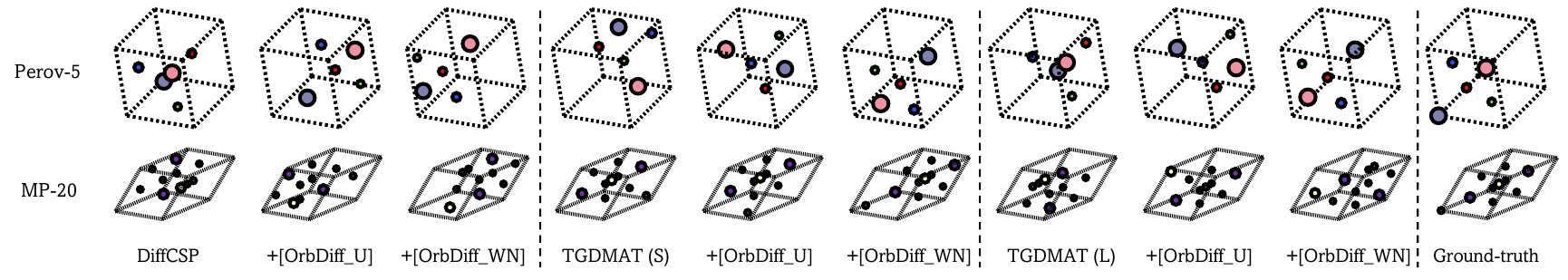}
    \caption{Qualitative comparison of Crystal Structure Predictions by 9 models, including DiffCSP, TGDMat (S) and TGDmat (L) with baselines, \texttt{OrbDiff\_U}, and \texttt{OrbDiff\_WN} against ground-truth samples on randomly selected samples from Perov-5 and MP-20 dataset.}
    \label{fig:diffcsp}
\end{figure}
\vspace{-0.3cm}

%% file: DELTA_sections/Experiments/table_diffcsp.tex
\begin{table}[htp]
\begin{minipage}{0.48\textwidth}
\centering
\caption{Text-guided CSP with TGDMat.}
\label{tab:tgdmat}
\setlength{\tabcolsep}{1.4pt}
\renewcommand{\arraystretch}{0.9}
\resizebox{0.95\textwidth}{!}{ 
\begin{tabular}{lcccc}
\toprule
\multirow{2}{*}{Method} & \multicolumn{2}{c}{Perov-5} & \multicolumn{2}{c}{MP-20} \\
\cmidrule(lr){2-3} \cmidrule(lr){4-5}
& Match (↑) & RMSE (↓) & Match (↑) & RMSE (↓) \\
\midrule
TGDMat (S)         &  59.39 & 0.066 & 59.90 & 0.078  \\
\rowcolor[HTML]{EAEAEA}
+ {[\texttt{OrbDiff\_U}]}                 &   63.51 &  0.062 & 56.50 & 0.085 \\
\rowcolor[HTML]{EAEAEA}
+ {[\texttt{OrbDiff\_WN}]}                 &\textbf{65.57}  & \textbf{0.054} &  \textbf{61.29} & \textbf{0.072} \\
\midrule
TGDMat (L)          & 95.17 & 0.013 &  61.91 &  0.081 \\
\rowcolor[HTML]{EAEAEA}
+ {[\texttt{OrbDiff\_U}]}                 &  95.88  & \textbf{0.012} & 65.94 & \textbf{0.069} \\
\rowcolor[HTML]{EAEAEA}
+ {[\texttt{OrbDiff\_WN}]}                 & \textbf{95.98} & \textbf{0.012} & \textbf{66.74} & \textbf{0.069}  \\
\bottomrule
\end{tabular}
}

\end{minipage}
\begin{minipage}{0.48\textwidth}

\caption{Crystal Structure Prediction (CSP). }
\label{tab:diffcsp}
\setlength{\tabcolsep}{1.4pt}
\renewcommand{\arraystretch}{1.027}
\resizebox{0.95\textwidth}{!}{ 
\begin{tabular}{lcccc}
\toprule
\multirow{2}{*}{Method} & \multicolumn{2}{c}{Perov-5} & \multicolumn{2}{c}{MP-20} \\
\cmidrule(lr){2-3} \cmidrule(lr){4-5}
& Match (↑) & RMSE (↓) & Match (↑) & RMSE (↓) \\
\midrule
P-cG-SchNet     & 48.22 & 0.418  & 15.39 & 0.376 \\
CDVAE            & 45.31 & 0.114  & 33.90 & 0.105  \\
\midrule
DiffCSP          & 52.02 & 0.076  & 51.49 & 0.063 \\
\rowcolor[HTML]{EAEAEA}
+ {[\texttt{OrbDiff\_U}]}  & 52.29&0.078  & 54.47 & 0.054 \\
\rowcolor[HTML]{EAEAEA}
+ {[\texttt{OrbDiff\_WN}]}    & \textbf{52.39} & \textbf{0.069} & \textbf{55.70} & \textbf{0.053}  \\
\bottomrule
\end{tabular}
}

\end{minipage}

\end{table}

%% file: DELTA_sections/Experiments/protein.tex
\subsection{Protein Structure Generation with \textsc{ProteiNA}} 
\label{exp:proteina}

\begin{figure}[H] 
    \centering
    \begin{minipage}[H]{0.65\linewidth}
        \centering
        \begin{table}[H] 

        \caption{Protein Structure Generation. Full comparisons are in the appendix. "+ [finetune]" denotes \(\mathcal{M}_{FS}^\text{no-tri}\) finetuned with the original loss; "+ [\texttt{OrbDiff}]" uses \texttt{OrbDiff} for finetuning. Full table can be found in~\cref{sub_app:PSG_full}}
        \vspace{0.5cm}
        \label{tab:protein-generation}
        \setlength{\tabcolsep}{1.4pt}
        \renewcommand{\arraystretch}{0.9}
        \resizebox{\linewidth}{!}{%
        \begin{tabular}{lccccccc}
        \toprule
        \multirow{2}{*}{Model} &  \multicolumn{2}{>{\columncolor[HTML]{f2cc8f}}c}{Designability} & \multicolumn{2}{c}{Diversity} & \multicolumn{2}{c}{Novelty} \\
        \cmidrule(lr){2-3} \cmidrule(lr){4-5} \cmidrule(lr){6-7}
         & Fraction~(↑) &scRMSD~(↓) & Cluster~(↑) & TM-score~(↓) & PDB~(↓) & AFDB~(↓) \\
        \midrule
        FoldFlow (OT) & 97.2 & - & 0.37 & 0.41 & 0.71 & 0.75 \\
        \(\mathcal{M}_\text{21M}\) & \textbf{99.0} & \textbf{0.72} & 0.30 & 0.39 & 0.81 & 0.84 \\
        \midrule
        \(\mathcal{M}^\text{no-tri}_\text{FS}\) & 93.8 & 1.04 & \textbf{0.62} & \textbf{0.36} & \textbf{0.69} & \textbf{0.76} \\
        + [finetune] & 93.8 & 1.00 & {0.54} & {0.37} & 0.74 & 0.83 \\
        \rowcolor[HTML]{EAEAEA}
        + [\texttt{OrbDiff}] & \textbf{95.6} & \textbf{0.93} & 0.52 & {0.37} & 0.74 & 0.83 \\
        \bottomrule
        \end{tabular}%
        }
        \end{table}
    \end{minipage}%
    \hfill
    \begin{minipage}[H]{0.34\linewidth}
        \centering
        \vspace{-0.05cm}
        \includegraphics[width=0.94\linewidth]{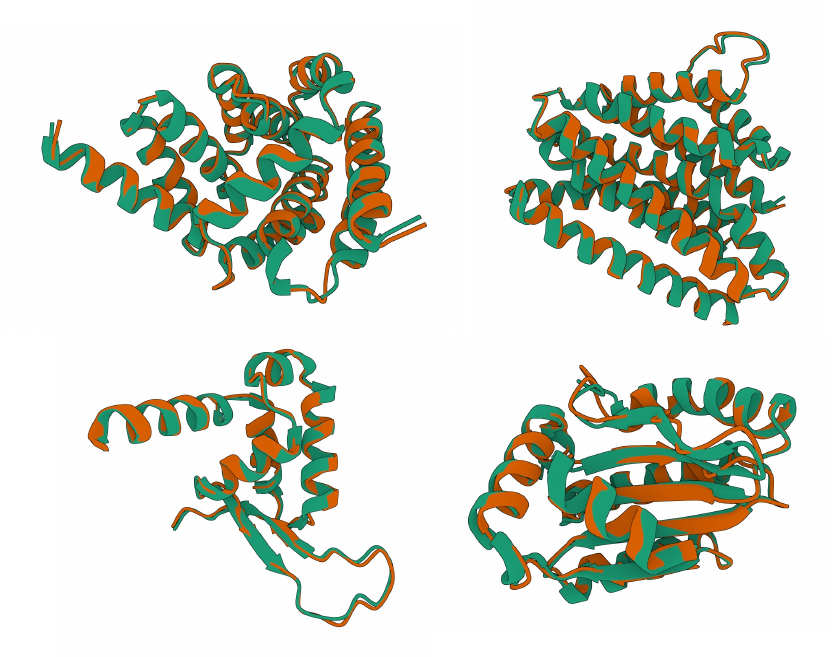}
        \vspace{-0.3cm}
        \caption{\texttt{OrbDiff} (orange) generated structures versus reference structure (green).}
        \label{fig:proteina_viz}
    \end{minipage}
\end{figure}
\vspace{-0.15cm}

\vspace{-0.3cm} 

We adopt \textsc{ProteiNA} (denoted \(\mathcal{M}\))~\citepcolor{proteina}, the state-of-the-art model for protein structure generation, as a backbone. Although \textsc{ProteiNA} is a non-equivariant transformer, it performs well through extensive data augmentation. We finetune the 200M-parameter version of \textsc{ProteiNA}—\(\mathcal{M}^\text{no-tri}_\text{FS}\)—using \texttt{OrbDiff}, applying Rao-Blackwellization with a uniform proposal distribution over $\mathsf{SO}(3)$, sampling 10,000 group elements.

For comparison, we also include the best-performing equivariant baseline, FoldFlow (OT)~\citepcolor{bose2023se}. We evaluate protein structures using three metrics: Designability (the feasibility of synthesizing the generated structures), Diversity, and Novelty. Designability is the most critical, while Diversity and Novelty are based on the designable samples, ensuring that the generated structures are synthetically plausible, diverse, and novel. Evaluation metric details are in~\cref{sub_app:PSG_eval}.

\cref{tab:protein-generation} shows that Orbit Diffusion boosts designability of $\mathcal{M}_\text{FS}^\text{no-tri}$ to 95.6\% and lowers scRMSD to 0.93, outperforming naive finetuning while maintaining competitive diversity (Cluster: 0.52) and novelty (PDB: 0.74). The state-of-the-art $\mathcal{M}_\text{21M}$ (400M parameters) achieves higher designability (99.0\%) and scRMSD (0.72) but at the cost of much lower diversity (Cluster: 0.30) and novelty (PDB: 0.81), showing a trade-off between validity and structural variety.

%% file: DELTA_sections/Experiments/analysis.tex
\subsection{Benefits of OrbDiff: Efficiency, Stability, and Equivariance}
\label{sec:train_eff}

\begin{figure}[!ht]
    \centering
    \includegraphics[]{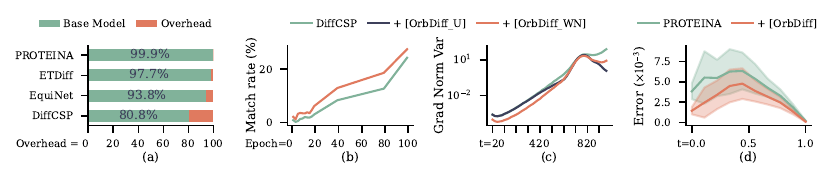}
    \caption{(a) \texttt{OrbDiff} introduces minimal computational overhead. (b) Match rate (\(\uparrow\)) during DiffCSP training on the Perov-5 dataset: \texttt{OrbDiff} accelerates convergence. (c) \texttt{OrbDiff} reduces gradient variance across noise levels. (d) \texttt{OrbDiff} improves the equivariance of \textsc{ProteiNA}.}
    \label{fig:merged_plot}
\end{figure}



Adding \texttt{OrbDiff} introduces minimal overhead (\cref{fig:merged_plot}a)—about 20\% for smaller models such as DiffCSP, and only 0.1\% for larger ones like \textsc{ProteiNA} (200M parameters). In a 24-hour training run, this corresponds to just 1.4 additional minutes. In practice, the extra memory and computation for sampling and averaging over group elements are negligible compared to the overall training cost. For example, in our \textsc{ProteiNA} experiments—one of the largest protein diffusion models—we use 10{,}000 SO(3) samples per training example, which adds only $\sim$40~MB per GPU to the total memory usage of $\sim$54~GB.

To better understand how \texttt{OrbDiff} accelerates convergence, we compare DiffCSP and DiffCSP + [\texttt{OrbDiff}] across training checkpoints. As shown in \cref{fig:merged_plot}b, integrating \texttt{OrbDiff} consistently improves match rates throughout training, especially in the early stages. 

We also compare the empirical gradient norm variance of DiffCSP (Perov-5) with \texttt{OrbDiff\_U} and \texttt{OrbDiff\_WN} across the full training set and various timesteps. As expected, \texttt{OrbDiff\_WN} achieves significantly lower variance at small to intermediate noise levels by sampling locally around \(x_0\), while \texttt{OrbDiff\_U} performs better at high noise levels due to its more global sampling. Both methods substantially reduce variance compared to DiffCSP. Finally, to assess equivariance preservation, we compare \textsc{ProteiNA} + [finetune] and \textsc{ProteiNA} + [\texttt{OrbDiff}] using an equivariance test from~\citepcolor{proteina}:
\begin{align}
\text{Error}_t = \mathbb{E}_{x \sim \hat{q}^G(x_0),\; x_t \sim \hat{q}_t^G(x_t \mid x_0),\; g \sim \text{Unif}(SO(3))} 
\left[ \text{RMSD}(g \circ \phi(x_t, t),\; \phi(g \circ x_t, t)) \right] .
\end{align}

As shown in \cref{fig:merged_plot}d, \texttt{OrbDiff} substantially reduces equivariance error compared to naive finetuning, indicating improved geometric consistency in the model’s denoising.

%% file: DELTA_sections/related_work.tex
\section{Related Work}



Equivariant neural networks have been extensively studied for their ability to encode symmetry priors in domains such as vision~\citepcolor{equiconv, harmonicnet}, 3D geometry~\citepcolor{tensorfieldnet, vectorneuron}, and molecular modeling~\citepcolor{bioref, scaledeepequi}. More recently, these ideas have been incorporated into diffusion models to improve generative performance in structured domains like proteins and molecules~\citepcolor{corso2024discovery, hoogeboom2022equivariant, igashov2024equivariant}. By designing the denoising network to be equivariant under a symmetry group, these models better align with the underlying data distribution. A comprehensive discussion of related work is provided in~\cref{app:related_work}.

%% file: DELTA_sections/conclusion.tex
\section{Conclusion}
\label{sec:conclusion}


Orbit Diffusion is a framework for training generative models under symmetry constraints by reducing gradient variance through Rao–Blackwellization. The approach unifies equivariant architectures and data augmentation strategies within a single probabilistic formulation, providing a provably lower-variance estimator while maintaining computational efficiency. Theoretically, we show that the proposed loss admits equivariant minimizers and connects to existing score-based and diffusion formulations. Empirically, Orbit Diffusion demonstrates strong and consistent performance across a diverse set of generative tasks, including molecular, crystal structure, and protein structure modeling. By bridging the gap between symmetry-aware modeling and optimization stability, our method improves both the scalability and practical applicability of equivariant generative models in scientific domains.

\section{Limitations}

To estimate the conditional expectation in our gradient estimator, we employ orbit sampling to reduce variance, which leads to improved performance. While efficient, our sampling scheme inherently constrains the choice of proposal distributions and may limit modeling flexibility, especially for tasks requiring diverse or adaptive noise structures. Furthermore, the number of orbit samples directly affects both computational cost and estimation accuracy, suggesting that dynamically adjusting the sampling strategy during training could yield better efficiency–accuracy trade-offs. Future work could explore adaptive or learned proposal mechanisms to enhance generalization and robustness across broader data regimes.

\section*{Acknowledgements}
This work was supported in part by the Deutsche Forschungsgemeinschaft (DFG, German Research Foundation) under Germany’s Excellence Strategy – EXC 2075 – 390740016; the DARPA ANSR program under award FA8750-23-2-0004; the DARPA CODORD program under award HR00112590089; the NSF grant IIS-1943641; the National University of Singapore Start-up Grant (Award No. SUG-251RES250); and gifts from Adobe Research, Cisco Research, and Amazon. We also acknowledge the support of the Stuttgart Center for Simulation Science (SimTech). VT and MN thank the International Max Planck Research School for Intelligent Systems (IMPRS-IS) for support. This work further received partial support from the Diabetes Center Berne and from the Ministry of Science, Research and the Arts Baden-Württemberg through the Artificial Intelligence Software Academy (AISA).

%% file: DELTA_sections/appendix.tex
\input{DELTA_sections/Appendix_sections/related_work}

\input{DELTA_sections/Appendix_sections/theoretical_results}

\input{DELTA_sections/Appendix_sections/experimental_details}

\input{DELTA_sections/Appendix_sections/variance_bias}

%% file: DELTA_sections/Appendix_sections/related_work.tex
\section{Related Work}
\label{app:related_work}

\paragraph{Equivariant Neural Networks.}  
Equivariant neural networks have attracted significant attention for tasks involving structured data, such as computer vision~\citepcolor{equiconv, harmonicnet}, 3D modeling~\citepcolor{tensorfieldnet, e3transformer, vectorneuron}, quantum mechanics and quantum field theory~\citepcolor{quantumfield, schrodinger}, biomolecular design~\citepcolor{bioref, scaledeepequi, proteina}. These networks exploit group symmetries to ensure consistent outputs under transformations such as rotations, translations, and permutations, which are commonly encountered in many scientific domains. By incorporating symmetric inductive biases, equivariant networks enhance model generalization and reduce data requirements by naturally encoding symmetry constraints. However, challenges remain, such as high computational complexity~\citepcolor{efficientequi} and difficulties in effectively learning with stochastic gradient descent (SGD)~\citepcolor{nonuniversalityequi}.
\paragraph{Equivariance and Diffusion Models.}  
Diffusion models have become a dominant class of generative models, known for their effectiveness in modeling complex data distributions~\citepcolor{song2020denoising, ho2020denoising, karras2022elucidating}. They work by gradually adding noise to data in a forward process and learning to reverse this corruption using a denoising network $\phi_\theta$. Incorporating symmetry, particularly equivariance, into these models has shown significant benefits in domains where data lies on geometric manifolds with known symmetries—such as structural biology~\citepcolor{corso2024discovery, yim2023se, schneuing2024structure, igashov2024equivariant}, molecular modeling~\citepcolor{hoogeboom2022equivariant, guan20233d, le2023navigating}, and material design~\citepcolor{diffcsp_bl2, diffcsp_bl1, jiao2023crystal}. In such settings, where symmetries are typically governed by the Euclidean group or its subgroups, enforcing equivariance in the generative process helps ensure physical plausibility and improves generalization. A common approach is to design the denoiser $\phi_\theta$ to be equivariant under a symmetry group \(G\), which encourages the learned distribution to converge toward the correct invariant target~\citepcolor{hoogeboom2022equivariant, geodiff, bose2023se, Theoequi}. 

Nonetheless, recent advances have shown that models without explicit equivariant constraints can still achieve strong empirical performance, thanks to greater flexibility in architectural design and effective use of data augmentation. These approaches may implicitly capture symmetry through training strategies rather than architectural bias, as demonstrated by state-of-the-art models such as \textsc{ProteiNA}~\citepcolor{proteina} and AlphaFold 3~\citepcolor{abramson2024accurate}.

%% file: DELTA_sections/Appendix_sections/theoretical_results.tex
\section{Theoretical Proofs}

\subsection{Proof of~\cref{the:theo_1}}
\label{sub_app:theo_1}

\begingroup
\setcounter{theorem}{0} 
\begin{tcolorbox}[title=Variance Reduction of Rao-Blackwell estimator.]
\begin{theorem}
Let \(\widehat{\nabla}_\phi [\mcL_t^G(\phi)]\) and \(\widehat{\nabla}_\phi^{(RB)} [\mcL_t^G(\phi)]\) denote the gradient estimators of \cref{eqn:sym_gradient} and \cref{eqn:new_sym_grad}, respectively. Suppose we can compute \( \mathbb{E}_{x_0 \sim \hat{q}^G_t(x_0 \mid x_t)}[x_0] \). Then
\[
        \emph{Var}\left(\widehat{\nabla}_\phi^{(RB)} [\mcL_t^G(\phi)]\right) \leq \emph{Var}\left(\widehat{\nabla}_\phi [\mcL_t^G(\phi)]\right).
\]
Moreover, the inequality is \emph{strict} unless \(\hat{q}^G_t(x_0 \mid x_t)\) is a Dirac delta, which is rarely the case in generative modeling where \(x_0\) is typically stochastic given \(x_t\).
\end{theorem}
\end{tcolorbox}
\endgroup

\begin{proof}
    
By definition of the Rao-Blackwellized estimator, we have
\[
    \widehat{\nabla}_\phi^{(RB)} [\mcL_t^G(\phi)] = \mathbb{E}_{x_0 \sim \hat{q}_t^G(x_0 \mid x_t)}\left[ \widehat{\nabla}_\phi [\mcL_t^G(\phi)] \mid x_t \right].
\]
Thus, the Rao-Blackwell theorem implies that conditioning reduces variance:
\[
    \operatorname{Var}\left( \mathbb{E}\left[ \widehat{\nabla}_\phi [\mcL_t^G(\phi)] \mid x_t \right] \right) \leq \operatorname{Var}\left( \widehat{\nabla}_\phi [\mcL_t^G(\phi)] \right),
\]
with equality if and only if \(\widehat{\nabla}_\phi [\mcL_t^G(\phi)]\) is almost surely a function of \(x_t\) (i.e., deterministic given \(x_t\)).

To verify this explicitly, consider the variance of the Rao-Blackwellized estimator:
\begin{align*}
    \operatorname{Var}\left(\widehat{\nabla}_\phi^{(RB)} [\mcL_t^G(\phi)]\right)
    &= \mathbb{E}_{x_t} \left[ \left( \mathbb{E}\left[ \widehat{\nabla}_\phi [\mcL_t^G(\phi)] \mid x_t \right] - \nabla_\phi [\mcL_t^G(\phi)] \right)^2 \right] \\
    &= \mathbb{E}_{x_t} \left[ \left( \mathbb{E}\left[ \widehat{\nabla}_\phi [\mcL_t^G(\phi)] - \nabla_\phi [\mcL_t^G(\phi)] \mid x_t \right] \right)^2 \right] \\
    &\leq \mathbb{E}_{x_t} \left[ \mathbb{E}\left[ \left( \widehat{\nabla}_\phi [\mcL_t^G(\phi)] - \nabla_\phi [\mcL_t^G(\phi)] \right)^2 \mid x_t \right] \right] \\
    &= \operatorname{Var}\left(\widehat{\nabla}_\phi [\mcL_t^G(\phi)]\right),
\end{align*}
where the inequality follows from Jensen’s inequality (applied to the convex function \(f(z) = z^2\)).

Therefore, the Rao-Blackwellized estimator has variance less than or equal to the original estimator, and strictly less unless the conditional variance given \(x_t\) is zero.
\end{proof}

\subsection{Equivalence of Symmetrized and Equivariant Diffusion Losses}
\label{sub_app:equivalance}

\begin{tcolorbox}
\begin{proposition}
Let \(\mcL_t^{(1)}(\phi)\) denote the \emph{Symmetrized Diffusion Loss}, defined as
\begin{align*}
    \mcL_t^{(1)}(\phi) = \mathbb{E}_{x'_0 \sim \hat{q}^G(x_0)} \mathbb{E}_{x_t \sim \hat{q}_t^G(x_t \mid x'_0)} \left[ \left\| \phi(x_t, t) - \mathbb{E}_{x_0 \sim \hat{q}_t^G(x_0 \mid x_t)}[x_0] \right\|^2 \right],
\end{align*}
and let \(\mcL_t^{(2)}(\phi)\) denote the corresponding loss defined on the original (non-symmetrized) data distribution \(\hat{q}(x_0)\):
\begin{align*}
    \mcL_t^{(2)}(\phi) = \mathbb{E}_{x'_0 \sim \hat{q}(x_0)} \mathbb{E}_{x_t \sim \hat{q}_t(x_t \mid x'_0)} \left[ \left\| \phi(x_t, t) - \mathbb{E}_{x_0 \sim \hat{q}_t^G(x_0 \mid x_t)}[x_0] \right\|^2 \right].
\end{align*}
Suppose \(\phi(x_t, t)\) is a \(G\)-equivariant function, i.e.,
\begin{align*}
    \phi(g \circ x_t, t) = g \circ \phi(x_t, t) \quad \forall\, g \in G,\; x_t \in \Omega.
\end{align*}
Then \(\mcL_t^{(1)}(\phi)\) and \(\mcL_t^{(2)}(\phi)\) are equivalent in the sense that they share the same minimizer and gradient with respect to \(\phi\).
\label{proposition_1}
\end{proposition}
\end{tcolorbox}

To prove this, we need the following lemma:
\begin{tcolorbox}[title=Symmetrized Forward Diffusion Distributions]
\begin{lemma}
Let \(\hat{q}(x_0)\) be an empirical distribution and \(\hat{q}^G(x_0)\) its symmetrized counterpart under a symmetry group \(G\), defined by \(\hat{q}^G(x_0) = S_G[\hat{q}](x_0)\). Suppose a forward diffusion process acting on \(\hat{q}(x_0)\) yields time-dependent marginal distributions \(\hat{q}_t(x_t)\). Let a similar process act on \(\hat{q}^G(x_0)\) to generate \(\hat{q}_t^G(x_t)\). Then, for all \(t \geq 0\), the following holds:
\[
\hat{q}_t^G(x_t) = S_G[\hat{q}_t](x_t).
\]
\label{lemma_1}
\end{lemma}
\end{tcolorbox}

The proof can be found in~\cref{sub_app:symmetrized}. 

We also need the following lemma:
\begin{tcolorbox}
\begin{lemma}
Let \( G \) be a group of isometries acting on \( \Omega \), and suppose the distribution \( \hat{q}_t^G(x_t \mid x_0) \)  is equivariant under the action of \( G \). Then for any \( g \in G \), the following identity holds:
\[
\mathbb{E}_{x_0 \sim \hat{q}_t^G(x_0 \mid g \circ x_t)}[x_0] = g \circ \mathbb{E}_{x_0 \sim \hat{q}_t^G(x_0 \mid x_t)}[x_0].
\]
\label{lemma_2}
\end{lemma}
\end{tcolorbox}

\begin{proof}
We start by expanding the conditional expectation:
\begin{align*}
    \mathbb{E}_{x_0 \sim \hat{q}_t^G(x_0 \mid g \circ x_t)}[x_0] 
    &= \int_\Omega x_0 \, \hat{q}_t^G(x_0 \mid g \circ x_t) \, \mathrm{d}x_0 \\
    &= \int_\Omega x_0 \, \hat{q}_t^G(g \circ x_t \mid x_0) \frac{\hat{q}^G(x_0)}{\hat{q}_t^G(g \circ x_t)} \, \mathrm{d}x_0.
\end{align*}

Now, apply the change of variable \(x_0 = g \circ \bar{x}_0\). Since \(g\) is an isometry, the Lebesgue measure is invariant, i.e., \(\mathrm{d}x_0 = \mathrm{d}\bar{x}_0\). Therefore:
\begin{align*}
    &= \int_\Omega (g \circ \bar{x}_0) \, \hat{q}_t^G(g \circ x_t \mid g \circ \bar{x}_0) \frac{\hat{q}^G(g \circ \bar{x}_0)}{\hat{q}_t^G(g \circ x_t)} \, \mathrm{d}\bar{x}_0 \\
    &= g \circ \int_\Omega \bar{x}_0 \, \hat{q}_t^G(x_t \mid \bar{x}_0) \frac{\hat{q}^G(\bar{x}_0)}{\hat{q}_t^G(x_t)} \, \mathrm{d}\bar{x}_0 \\
    &= g \circ \mathbb{E}_{x_0 \sim \hat{q}_t^G(x_0 \mid x_t)}[x_0],
\end{align*}
where the second equality follows from the equivariance of \(\hat{q}_t^G\) and the invariance of \(\hat{q}^G\) under the group action. This concludes the proof.
\end{proof}

Now the prove the~\cref{proposition_1}.

\begin{proof}
We begin by simplifying \(\mcL_t^{(1)}\):
\begin{align*}
    \mcL_t^{(1)}(\phi) &= \mathbb{E}_{x_t \sim \hat{q}_t^G(x_t)} \mathbb{E}_{x'_0 \sim \hat{q}^G(x_0 \mid x_t)} \left[ \left\| \phi(x_t, t) - \mathbb{E}_{x_0 \sim \hat{q}_t^G(x_0 \mid x_t)}[x_0] \right\|^2 \right] \\
    &= \mathbb{E}_{x_t \sim \hat{q}_t^G(x_t)} \left[ \left\| \phi(x_t, t) - \mathbb{E}_{x_0 \sim \hat{q}_t^G(x_0 \mid x_t)}[x_0] \right\|^2 \right].
\end{align*}
Similarly,
\begin{align*}
    \mcL_t^{(2)}(\phi) = \mathbb{E}_{x_t \sim \hat{q}_t(x_t)} \left[ \left\| \phi(x_t, t) - \mathbb{E}_{x_0 \sim \hat{q}_t^G(x_0 \mid x_t)}[x_0] \right\|^2 \right].
\end{align*}

We now rewrite \(\mcL_t^{(1)}\) in integral form:
\begin{align*}
    \mcL_t^{(1)}(\phi) = \int_{\Omega} \left\| \phi(x_t, t) - \mathbb{E}_{x_0 \sim \hat{q}_t^G(x_0 \mid x_t)}[x_0] \right\|^2 \hat{q}_t^G(x_t) \, \mathrm{d}x_t.
\end{align*}

Using the decomposition of the measure over the group action, we change variables:
\[
\int_\Omega f(x_t) \, \mathrm{d}x_t = \int_{\Omega/G} \int_G f(g \circ x_t) \, \mathrm{d}\mu_G(g) \, \mathrm{d}x_t.
\]

Thus,
\begin{align*}
\mcL_t^{(1)}(\phi) &= \int_{\Omega/G} \int_G \left\| \phi(g \circ x_t, t) - \mathbb{E}_{x_0 \sim \hat{q}_t^G(x_0 \mid g \circ x_t)}[x_0] \right\|^2 \hat{q}_t^G(g \circ x_t) \, \mathrm{d}\mu_G(g) \, \mathrm{d}x_t \\
&= \int_{\Omega/G} \int_G \left\| g \circ \phi(x_t, t) - \mathbb{E}_{x_0 \sim \hat{q}_t^G(x_0 \mid g \circ x_t)}[x_0] \right\|^2 \hat{q}_t^G(x_t) \, \mathrm{d}\mu_G(g) \, \mathrm{d}x_t,
\end{align*}
where we used the \(G\)-equivariance of \(\phi\) and the invariance of \(\hat{q}_t^G\).

Since \(G\) acts isometrically, we apply:
\begin{align*}
\left\| g \circ \phi(x_t, t) - \mathbb{E}_{x_0 \sim \hat{q}_t^G(x_0 \mid g \circ x_t)}[x_0] \right\|^2 
= \left\| \phi(x_t, t) - g^{-1} \circ \mathbb{E}_{x_0 \sim \hat{q}_t^G(x_0 \mid g \circ x_t)}[x_0] \right\|^2.
\end{align*}

Next, applying \cref{lemma_2}, we have:
\begin{align*}
    \left\| g \circ \phi(x_t, t) - \mathbb{E}_{x_0 \sim \hat{q}_t^G(x_0 \mid g \circ x_t)}[x_0] \right\|^2  = \left\| \phi(x_t, t) - \mathbb{E}_{x_0 \sim \hat{q}_t^G(x_0 \mid x_t)}[x_0] \right\|^2 
\end{align*}

We conclude:
\begin{align*}
\mcL_t^{(1)}(\phi) &= \int_{\Omega/G} \int_G \left\| \phi(x_t, t) - \mathbb{E}_{x_0 \sim \hat{q}_t^G(x_0 \mid x_t)}[x_0] \right\|^2 \hat{q}_t^G(x_t) \, \mathrm{d}\mu_G(g) \, \mathrm{d}x_t \\
&= \int_{\Omega/G} \left\| \phi(x_t, t) - \mathbb{E}_{x_0 \sim \hat{q}_t^G(x_0 \mid x_t)}[x_0] \right\|^2 \hat{q}_t^G(x_t) \, \mathrm{d}x_t.
\end{align*}

Next, we similarly transform \(\mcL_t^{(2)}(\phi)\):
\begin{align*}
\mcL_t^{(2)}(\phi) &= \int_{\Omega/G} \int_G \left\| \phi(g \circ x_t, t) - \mathbb{E}_{x_0 \sim \hat{q}_t^G(x_0 \mid g \circ x_t)}[x_0] \right\|^2 \hat{q}_t(g \circ x_t) \, \mathrm{d}\mu_G(g) \, \mathrm{d}x_t \\
&= \int_{\Omega/G} \left\| \phi(x_t, t) - \mathbb{E}_{x_0 \sim \hat{q}_t^G(x_0 \mid x_t)}[x_0] \right\|^2 \left( \int_G \hat{q}_t(g \circ x_t) \, \mathrm{d}\mu_G(g) \right) \mathrm{d}x_t.
\end{align*}

Using the theoretical result from~\cref{sub_app:symmetrized}, we have:
\[
\hat{q}_t^G(x_t) = S_G[\hat{q}_t](x_t) = \int_G \hat{q}_t(g \circ x_t) \, \mathrm{d}\mu_G(g),
\]
we conclude:
\begin{align*}
\mcL_t^{(2)}(\phi) = \int_{\Omega/G} \left\| \phi(x_t, t) - \mathbb{E}_{x_0 \sim \hat{q}_t^G(x_0 \mid x_t)}[x_0] \right\|^2 \hat{q}_t^G(x_t) \, \mathrm{d}x_t = \mcL_t^{(1)}(\phi).
\end{align*}

Hence, the two loss functions are equivalent in the sense that they yield the same gradients and minimizers with respect to \(\phi\).
\end{proof}

\subsection{Symmetrized Forward Diffusion Distribution.}
\label{sub_app:symmetrized}

Below is the formal lemma and the proof for the symmetrized forward diffusion distribution. 
\begingroup
\setcounter{lemma}{0} 
\begin{tcolorbox}[title=Symmetrized Forward Diffusion Distributions]
\begin{lemma}
Let \(\hat{q}(x_0)\) be an empirical distribution and \(\hat{q}^G(x_0)\) its symmetrized counterpart under a symmetry group \(G\), defined by \(\hat{q}^G(x_0) = S_G[\hat{q}](x_0)\). Suppose a forward diffusion process acting on \(\hat{q}(x_0)\) yields time-dependent marginal distributions \(\hat{q}_t(x_t)\). Let a similar process act on \(\hat{q}^G(x_0)\) to generate \(\hat{q}_t^G(x_t)\). Then, for all \(t \geq 0\), the following holds:
\[
\hat{q}_t^G(x_t) = S_G[\hat{q}_t](x_t).
\]
\end{lemma}
\end{tcolorbox}
\endgroup

\begin{proof}
The marginal distribution at time step \(t\) of a diffusion process is defined as
\[
\hat{q}_t(x_t) = \int_\Omega q_t(x_t \mid x_0) \hat{q}(x_0) \, \mathrm{d}x_0,
\]
where \(q_t(x_t \mid x_0) = \mathcal{N}(x_t; \alpha_t x_0, \sigma_t^2 I)\) is the Gaussian diffusion kernel, which is equivariant under isometry group transformations. Specifically, for any \(g \in G\), we have
\[
q_t(x_t \mid x_0) = q_t(g \circ x_t \mid g \circ x_0).
\]
The symmetrized marginal distribution at time \(t\) is defined as
\begin{align*}
S_G[\hat{q}_t](x_t) &= \int_G \hat{q}_t(g \circ x_t) \mathrm{d}\mu_G(g) \\
&= \int_G \left[ \int_\Omega q_t(g \circ x_t \mid x_0) \hat{q}(x_0) \, \mathrm{d}x_0 \right] \mathrm{d}\mu_G(g) \\
&= \int_G \int_\Omega q_t(g \circ x_t \mid x_0) \hat{q}(x_0) \, \mathrm{d}x_0 \mathrm{d}\mu_G(g).
\end{align*}

Next, we compute the marginal distribution at time \(t\) when the forward process is applied to the symmetrized data distribution:
\begin{align*}
\hat{q}_t^G(x_t) &= \int_\Omega q_t(x_t \mid x_0) S_G[\hat{q}](x_0) \, \mathrm{d}x_0 \\
&= \int_\Omega q_t(x_t \mid x_0) \int_G \hat{q}(g \circ x_0) \mathrm{d}\mu_G(g) \mathrm{d}x_0 \\
&= \int_\Omega \int_G q_t(x_t \mid x_0) \hat{q}(g \circ x_0) \mathrm{d}\mu_G(g) \mathrm{d}x_0 \\
&= \int_G \int_\Omega q_t(x_t \mid x_0) \hat{q}(g \circ x_0) \mathrm{d}x_0 \mathrm{d}\mu_G(g).
\end{align*}

Applying a change of variable \(x_0 \mapsto g^{-1} \circ x_0\), we get
\begin{align*}
\hat{q}_t^G(x_t) &= \int_G \int_\Omega q_t(x_t \mid g^{-1} \circ x_0) \hat{q}(g \circ [g^{-1} \circ x_0]) \mathrm{d}(g^{-1} \circ x_0) \mathrm{d}\mu_G(g) \\
&= \int_G \int_\Omega q_t(x_t \mid g^{-1} \circ x_0) \hat{q}(x_0) \mathrm{d}x_0 \mathrm{d}\mu_G(g) \quad \text{(since the Jacobian of \(g\) is 1)} \\
&= \int_G \int_\Omega q_t(g \circ x_t \mid x_0) \hat{q}(x_0) \mathrm{d}x_0 \mathrm{d}\mu_G(g) \quad \text{(by kernel equivariance)}.
\end{align*}
Thus, we have
\[
\hat{q}_t^G(x_t) = S_G[\hat{q}_t](x_t).
\]
This completes the proof.
\end{proof}

\subsection{Proof of~\cref{theo_2}}
\label{sub_app:orbit_expectation}

\begin{tcolorbox}[title=Unbiased gradient and equivariance of \texttt{OrbDiff}]
\begin{theorem}
Let \( G \) be a locally compact isometry group acting on data space \( \Omega \), and suppose the forward kernels \( \hat{q}_t^G(x_t \mid x_0) \) are \( G \)-invariant: \( \hat{q}_t^G(g \circ x_t \mid g \circ x_0) = \hat{q}_t^G(x_t \mid x_0) \) for all \( g \in G \). Then:

\begin{enumerate}
    \item The \texttt{OrbDiff} target
\(
\phi^*(x_0, x_t, t)
\)
satisfies: \( \phi^*(x_0, h \circ x_t, t) = h \circ \phi^*(x_0, x_t, t) \) for all \( h \in G \).

    \item The gradient of the \texttt{OrbDiff} loss \eqref{eqn:final_sym_loss} equals that of the ideal loss \eqref{eqn:new_sym_grad}, i.e., \texttt{OrbDiff} provides an unbiased gradient estimator.
\end{enumerate}
\end{theorem}
\end{tcolorbox}

\textbf{We first prove that the \texttt{OrbDiff} target is equivariant:
}
\begin{proof}

We compute \( \phi^*(x_0, h \circ x_t, t) \) using the definition:
\[
\phi^*(x_0, h \circ x_t, t) = \frac{1}{Z(h \circ x_t, x_0)} \int_G (g \circ x_0)\, \hat{q}_t^G(h \circ x_t \mid g \circ x_0)\, \mathrm{d}\mu_G(g).
\]
By the equivariance of \( \hat{q}_t^G \), we have:
\[
\hat{q}_t^G(h \circ x_t \mid g \circ x_0) = \hat{q}_t^G(x_t \mid h^{-1} \circ g \circ x_0).
\]
Letting \( g' = h^{-1} \circ g \), so \( g = h \circ g' \), and using the left-invariance of the Haar measure \( \mu_G \), we get:
\[
\phi^*(x_0, h \circ x_t, t) = \frac{1}{Z(h \circ x_t, x_0)} \int_G (h \circ g' \circ x_0)\, \hat{q}_t^G(x_t \mid g' \circ x_0)\, \mathrm{d}\mu_G(g').
\]
Factoring out \( h \) from the integrand gives:
\[
= h \circ \left[ \frac{1}{Z(h \circ x_t, x_0)} \int_G (g' \circ x_0)\, \hat{q}_t^G(x_t \mid g' \circ x_0)\, \mathrm{d}\mu_G(g') \right].
\]
It remains to show that \( Z(h \circ x_t, x_0) = Z(x_t, x_0) \), where:
\[
Z(x_t, x_0) := \int_G \hat{q}_t^G(x_t \mid g \circ x_0)\, \mathrm{d}\mu_G(g).
\]
Using the same substitution:
\[
Z(h \circ x_t, x_0) = \int_G \hat{q}_t^G(h \circ x_t \mid g \circ x_0)\, \mathrm{d}\mu_G(g) 
= \int_G \hat{q}_t^G(x_t \mid h^{-1} \circ g \circ x_0)\, \mathrm{d}\mu_G(g)
= \int_G \hat{q}_t^G(x_t \mid g' \circ x_0)\, \mathrm{d}\mu_G(g') = Z(x_t).
\]
Therefore,
\[
\phi^*(x_0, h \circ x_t, t) = h \circ \phi^*(x_0, x_t, t).
\]
\end{proof}

\textbf{Next, we prove that \texttt{OrbDiff} yields an unbiased gradient estimator:} 

\begin{proof}

Our goal is to estimate the following gradient:
\begin{align}
    \nabla_\phi \mathcal{L}_t^{G}(\phi)
    = 2\mathbb{E}_{x_t \sim \hat{q}_t^G(x_t)} \left[  \phi(x_t, t) - \mathbb{E}_{x_0 \sim \hat{q}_t^G(x_0 \mid x_t)} [x_0]  \right].
\end{align}

Our proposed Rao-Blackwell loss function has the same gradient as \(\mathcal{L}_t^G(\phi)\), but with reduced variance due to the use of the conditional expectation \(\mathbb{E}_{x_0 \sim \hat{q}_t^G(x_0 \mid x_t)}[x_0]\):
\begin{align}
    \mathcal{L}_t^{\mathrm{RB}}(\phi) = \mathbb{E}_{x_0' \sim \hat{q}^G(x'_0)} \mathbb{E}_{x_t \sim \hat{q}_t^G(x_t \mid x_0')} \left[ \left\| \phi(x_t, t) - \mathbb{E}_{x_0 \sim \hat{q}_t^G(x_0 \mid x_t)}[x_0] \right\|^2 \right].
    \label{eqn:orbloss}
\end{align}

However, computing \(\mathbb{E}_{x_0 \sim \hat{q}_t^G(x_0 \mid x_t)} [x_0]\) can be computationally expensive. To address this, we introduce OrbDiff, which uses a biased proposal distribution to approximate this expectation using only samples from the orbit of the \(x_0\) that generated \(x_t\). This yields the alternative target:
\begin{align}
    \phi^*(x_0, x_t, t) = \frac{1}{Z(x_0, x_t)} \int_{G} (g \circ x_0) \, \hat{q}_t^G(g \circ x_0 \mid x_t) \, \mathrm{d}\mu_G(g),
\end{align}
where the normalization constant is
\begin{align}
    Z(x_0, x_t) = \int_{G} \hat{q}_t^G(g \circ x_0 \mid x_t) \, \mathrm{d}\mu_G(g).
\end{align}
This matches \cref{eqn:weight_orbit_integral}, which can be verified via straightforward transformations.

Although \(\phi^*\) differs from \(\mathbb{E}_{x_0 \sim \hat{q}_t^G(x_0 \mid x_t)} [x_0]\), we show that replacing the Rao-Blackwell target in Eq.~\eqref{eqn:orbloss} with \(\phi^*\) yields a loss whose gradient still matches the original gradient. This relies on the assumption that the forward conditional distribution is equivariant under the group action, i.e.,
\begin{align}
    \hat{q}_t^G(g \circ x_t \mid g \circ x_0) = \hat{q}_t^G(x_t \mid x_0), \quad \forall g \in G,
\end{align}
which is a natural condition satisfied in many generative models such as diffusion models or flow matching with isotropic Gaussian priors.

Under this assumption, the OrbDiff loss is given by:
\begin{align}
    \mathcal{L}_t^{\mathrm{OrbDiff}}(\phi)
    &= \mathbb{E}_{x_0 \sim \hat{q}^G(x_0)} \mathbb{E}_{x_t \sim \hat{q}_t^G(x_t \mid x_0)} \left[ \left\| \phi(x_t, t) - \phi^*(x_0, x_t, t) \right\|^2 \right] \\
    &= \mathbb{E}_{x_t \sim \hat{q}_t^G(x_t)} \mathbb{E}_{x_0 \sim \hat{q}_t^G(x_0 \mid x_t)} \left[ \left\| \phi(x_t, t) - \phi^*(x_0, x_t, t) \right\|^2 \right].
    \label{eqn:orb}
\end{align}

Taking the gradient with respect to \(\phi\), we obtain:
\begin{align}
    \nabla_\phi \mathcal{L}_t^{\mathrm{OrbDiff}}(\phi)
    &= 2 \mathbb{E}_{x_t \sim \hat{q}_t^G(x_t)} \mathbb{E}_{x_0 \sim \hat{q}_t^G(x_0 \mid x_t)} \left[ \phi(x_t, t) - \phi^*(x_0, x_t, t) \right] \\
    &= 2 \mathbb{E}_{x_t \sim \hat{q}_t^G(x_t)} \left[ \phi(x_t, t) - \mathbb{E}_{x_0 \sim \hat{q}_t^G(x_0 \mid x_t)} [\phi^*(x_0, x_t, t)] \right].
\end{align}

Thus, to ensure that OrbDiff yields the correct gradient, it suffices to show:
\begin{align}
    \mathbb{E}_{x_0 \sim \hat{q}_t^G(x_0 \mid x_t)} [\phi^*(x_0, x_t, t)]
    = \mathbb{E}_{x_0 \sim \hat{q}_t^G(x_0 \mid x_t)} [x_0].
\end{align}

We compute:
\begin{align}
    \mathbb{E}_{x_0 \sim \hat{q}_t^G(x_0 \mid x_t)} [\phi^*(x_0, x_t, t)]
    &= \int_\Omega \int_G \phi^*(g' \circ x_0, x_t, t) \, \hat{q}_t^G(g' \circ x_0 \mid x_t) \, \mathrm{d}\mu_G(g') \, \mathrm{d}x_0.
\end{align}

Substituting the expression for \(\phi^*\) and applying the change of variables \(g \mapsto g \cdot g'\) with left-invariant Haar measure \(\mu_G\), we have:
\begin{align}
    &= \int_\Omega \int_G \left[ \frac{1}{Z(g' \circ x_0, x_t)} \int_G (g \circ [g' \circ x_0]) \, \hat{q}_t^G(g \circ [g' \circ x_0] \mid x_t) \, \mathrm{d}\mu_G(g) \right] \hat{q}_t^G(g' \circ x_0 \mid x_t) \, \mathrm{d}\mu_G(g') \, \mathrm{d}x_0 \\
    &= \int_\Omega \int_G \left[ \frac{1}{Z(x_0, x_t)} \int_G (g \circ x_0) \, \hat{q}_t^G(g \circ x_0 \mid x_t) \, \mathrm{d}\mu_G(g) \right] \hat{q}_t^G(g' \circ x_0 \mid x_t) \, \mathrm{d}\mu_G(g') \, \mathrm{d}x_0 \\
    &= \int_\Omega \left[ \frac{1}{Z(x_0, x_t)} \int_G (g \circ x_0) \, \hat{q}_t^G(g \circ x_0 \mid x_t) \, \mathrm{d}\mu_G(g) \right] \left[ \int_G \hat{q}_t^G(g' \circ x_0 \mid x_t) \, \mathrm{d}\mu_G(g') \right] \mathrm{d}x_0 \\
    &= \int_\Omega \left[ \frac{1}{Z(x_0, x_t)} \int_G (g \circ x_0) \, \hat{q}_t^G(g \circ x_0 \mid x_t) \, \mathrm{d}\mu_G(g) \right] Z(x_0, x_t) \, \mathrm{d}x_0 \\
    &= \int_\Omega \int_G (g \circ x_0) \, \hat{q}_t^G(g \circ x_0 \mid x_t) \, \mathrm{d}\mu_G(g) \, \mathrm{d}x_0 \\
    &= \mathbb{E}_{x_0 \sim \hat{q}_t^G(x_0 \mid x_t)} [x_0].
\end{align}

Therefore, despite \(\phi^*\) not being equal to the conditional expectation at each \(x_t\), the gradient induced by the OrbDiff loss matches the desired gradient. OrbDiff thus provides an unbiased estimate of \(\nabla_\phi \mathcal{L}_t^{G}(\phi)\), while using only samples from the orbit of \(x_0\).

\end{proof}

\subsection{Unconstrained Non-Symmetrized Diffusion Minimizer is not guaranteed G-equivaraint}
\label{sub_app:non_sym_diff_mini}

\begin{tcolorbox}
\begin{lemma}
Let 
\[
    \mathcal{L}_t(\phi) = \mathbb{E}_{(x_0, x_t) \sim \hat{q}(x_0, x_t)} \left[ \| \phi(x_t, t) - x_0 \|^2 \right]
\]
be the unconstrained Non-Symmetrized Diffusion loss, where \(\hat{q}(x_0, x_t)\) is the empirical joint distribution of clean and noisy data. Then the minimizer \(\phi^*(x_t, t)\) of \(\mathcal{L}_t\) is the conditional expectation:
\[
    \phi^*(x_t, t) = \mathbb{E}_{x_0 \sim \hat{q}_t(x_0 \mid x_t)}[x_0].
\]
However, this minimizer is not guaranteed to be equivariant under the action of a symmetry group \(\Ge\).
\end{lemma}
\end{tcolorbox}

\begin{proof}
To find the minimizer of the diffusion loss, we first compute the stationary point. The loss is given by:
\[
    \mathcal{L}_t(\phi) = \mathbb{E}_{(x_0, x_t) \sim \hat{q}(x_0, x_t)} \left[ \| \phi(x_t, t) - x_0 \|^2 \right] 
    = \mathbb{E}_{x_t \sim \hat{q}_t(x_t)} \mathbb{E}_{x_0 \sim \hat{q}_t(x_0 \mid x_t)} \left[ \| \phi(x_t, t) - x_0 \|^2 \right].
\]
The gradient of the internal expectation is:
\begin{align*}
    \nabla_\phi \mathbb{E}_{x_0 \sim \hat{q}_t(x_0 \mid x_t)} \left[ \| \phi(x_t, t) - x_0 \|^2 \right] 
    &= 2 \mathbb{E}_{x_0 \sim \hat{q}_t(x_0 \mid x_t)} \left[ \phi(x_t, t) - x_0 \right] \\
    &= 2 \left( \phi(x_t, t) - \mathbb{E}_{x_0 \sim \hat{q}_t(x_0 \mid x_t)}[x_0] \right).
\end{align*}
Setting the gradient to zero gives the minimizer:
\[
    \phi^*(x_t, t) = \mathbb{E}_{x_0 \sim \hat{q}_t(x_0 \mid x_t)}[x_0] 
    = \int_\Omega x_0 \hat{q}_t(x_0 \mid x_t) \, dx_0.
\]

Next, we provide a counterexample to show that \(\phi^*(x_t, t)\) is not guaranteed to be equivariant.

\paragraph{Counterexample: Translation in 1D}
\begin{enumerate}
    \item \textbf{Data}: Two points \(x_0^1 = 0\) and \(x_0^2 = 1\), with uniform empirical distribution \(\hat{q}(x_0^i) = 0.5\).
    \item \textbf{Group action}: Translation by \(a = 1\), i.e., \(g \circ x = x + 1\).
    \item \textbf{Diffusion kernel}: \({q}_t(x_t \mid x_0) = \mathcal{N}(x_t; \alpha_t x_0, \sigma_t^2)\).
\end{enumerate}

First, rewrite the minimizer:
\begin{align*}
    \phi^*(x_t, t) &= \sum_{i=1}^N x_0^i \hat{q}_t(x_0^i \mid x_t) \\
    &= \sum_{i=1}^N x_0^i \hat{q}_t(x_t \mid x_0^i) \frac{\hat{q}(x_0^i)}{\hat{q}_t(x_t)} \\
    &= \frac{1}{\hat{q}_t(x_t)} \sum_{i=1}^N x_0^i \hat{q}_t(x_t \mid x_0^i) \hat{q}(x_0^i).
\end{align*}
Substituting \(x_0^1 = 0\) and \(x_0^2 = 1\):
\begin{align*}
    \phi^*(x_t, t) & = \frac{1}{\hat{q}_t(x_t)} \left( 0 \cdot \mathcal{N}(x_t; 0, \sigma_t^2) \cdot 0.5 + 1 \cdot \mathcal{N}(x_t; \alpha_t, \sigma_t^2) \cdot 0.5 \right), \\ 
    & = \frac{0.5 \cdot \mathcal{N}(x_t; \alpha_t, \sigma_t^2)}{\hat{q}_t(x_t)} 
    = \frac{\mathcal{N}(x_t; \alpha_t, \sigma_t^2)}{\mathcal{N}(x_t; 0, \sigma_t^2) + \mathcal{N}(x_t; \alpha_t, \sigma_t^2)}.
\end{align*}
Now, compute \(\phi^*(g \circ x_t, t)\) by applying \(g \circ x_t = x_t + 1\):
\[
    \phi^*(g \circ x_t, t) = \frac{\mathcal{N}(x_t + 1; \alpha_t, \sigma_t^2)}{\mathcal{N}(x_t + 1; 0, \sigma_t^2) + \mathcal{N}(x_t + 1; \alpha_t, \sigma_t^2)} < 1.
\]
Next, compute \(g \circ \phi^*(x_t, t)\):
\[
    g \circ \phi^*(x_t, t) = \frac{\mathcal{N}(x_t; \alpha_t, \sigma_t^2)}{\mathcal{N}(x_t; 0, \sigma_t^2) + \mathcal{N}(x_t; \alpha_t, \sigma_t^2)} + 1 > 1.
\]
Thus, 
\[
\phi^*(g \circ x_t, t) < g \circ \phi^*(x_t, t),
\]
since \(\phi^*(g \circ x_t, t) < 1\) and \(g \circ \phi^*(x_t, t) > 1\). Consequently,
\[
\phi^*(g \circ x_t, t) \neq g \circ \phi^*(x_t, t).
\]
This completes the counterexample, showing that \(\phi^*(x_t, t)\) is not necessarily equivariant.
\end{proof}

%% file: DELTA_sections/Appendix_sections/experimental_details.tex
\section{Experimental Details}
\label{app:exp_details}

\subsection{Molecular Conformer Generation (MCG) with Flow Matching}
\label{sup_app:mcg}

Molecular conformer generation is a fundamental task in computational chemistry and drug discovery, where the goal is to generate plausible 3D structures (conformers) that correspond to a 2D molecular graph. Molecular conformer generation is essential for drug discovery and molecular property prediction, as the 3D structure greatly influences chemical behavior and interactions~\citepcolor{Liu2022MolecularGPA, Axelrod2020MolecularMLA}. 

To model this task effectively, it is crucial to respect the underlying symmetries of molecular structures. Since conformers are invariant under global rotations and translations, one might consider the full Euclidean group \(\mathsf{SE}(3)\). However, in practice, molecular structures are typically zero-centered, effectively removing the need to model translation invariance. As a result, it suffices to consider equivariance under the rotation group \(\mathsf{SO}(3)\). In addition, the molecular graph may exhibit symmetry under automorphisms—permutations of atoms that preserve the graph structure—making it important to account for graph isomorphism to avoid redundant representations and ensure physically meaningful predictions.

\subsubsection{MCG - Dataset}

We evaluate our method on the \textsc{GEOM-QM9} dataset~\citepcolor{axelrod2022geom}, a widely used subset of \textsc{GEOM} containing molecules with an average of 11 atoms. We follow the same train/validation/test split as in~\citepcolor{geomol, jing2022torsional}, consisting of 106,586 / 13,323 / 1,000 molecules, respectively.

\subsubsection{MCG - Baselines}
\label{sub_app:QM9_baseline}

We compare against strong recent baselines with publicly available code, including \textsc{GeoDiff}~\citepcolor{xu2022geodiff}, \textsc{GeoMol}~\citepcolor{geomol}, Torsional Diffusion~\citepcolor{jing2022torsional}, \textsc{MCF}~\citepcolor{Swallow}, and \textsc{ETFlow}~\citepcolor{hassan2024flow}. \textsc{GeoDiff} generates structures using a roto-translationally invariant diffusion process, starting from an invariant initial density and evolving through a Markov kernel that preserves this invariance. \textsc{GeoMol} predicts 3D structures by modeling torsion angles conditioned on a molecular graph, offering fast inference and good geometric validity. Torsional Diffusion generates conformers using a diffusion process in torsion space. \textsc{MCF} directly models 3D coordinates using a diffusion model without enforcing equivariance, relying instead on model scale to achieve strong performance. \textsc{ETFlow}, the strongest of these baselines, is an equivariant flow matching model that uses a harmonic prior to encourage spatial proximity of bonded atoms. We integrate Orbit Diffusion into \textsc{ETFlow} to build on its strong geometric foundation. 

\subsubsection{MCG - \textsc{ETFlow} with Orbit Diffusion}
\label{sub_app:orb_flow}

While Orbit Diffusion is framed within the context of diffusion models, it naturally extends to flow matching. We introduce this extension through the design of \textsc{ETFlow}, which employs a harmonic prior and a flexible coupling to the data distribution. Unlike diffusion models, which fix the prior and reverse-time coupling, flow matching allows arbitrary choices for both~\citepcolor{tong2023improving}, offering greater flexibility in the generative process.

Assume a coupling \( q(x_0, x_1) \) between the base distribution \( q_0 \) and the data distribution \( q_1 \). For each pair \( (x_0, x_1) \), define the linear interpolation:
\[
I_t(x_0, x_1) = (1 - t)x_0 + t x_1, \quad t \in [0, 1].
\]

\textbf{Note:} In contrast to diffusion models (where \( t = 0 \) corresponds to data and \( t = 1 \) to noise), flow matching treats \( x_0 \sim q_0 \) as the prior sample and \( x_1 \sim q_1 \) as the data sample.

\textsc{ETFlow} defines the conditional distribution:
\[
q_t(x_t \mid x_0, x_1) = \mathcal{N}\left(x_t \mid I_t(x_0, x_1), \sigma^2 t(1 - t)\right),
\]
with small \(\sigma\), inducing the following velocity field:
\[
v_t(x_t) = x_1 - x_0 + \frac{1 - 2t}{2 \sqrt{t(1 - t)}} \epsilon, \quad \epsilon \sim \mathcal{N}(0, I).
\]

Given the sampling equation:
\[
x_t = (1 - t)x_0 + t x_1 + \sigma \sqrt{t(1 - t)} \epsilon,
\]
we can express \(\epsilon\) as:
\[
\epsilon = \frac{x_t - (1 - t)x_0 - t x_1}{\sigma \sqrt{t(1 - t)}}.
\]

Substituting this into \(v_t(x_t)\) yields:
\begin{align*}
v_t(x_t) &= x_1 - x_0 + \frac{1 - 2t}{2\sigma t(1 - t)} \left( x_t - (1 - t)x_0 - t x_1 \right) \\
&= \frac{1 - 2t}{2\sigma t(1 - t)} x_t - \left(1 + \frac{1 - 2t}{2\sigma t}\right)x_0 + \left(1 - \frac{1 - 2t}{2\sigma(1 - t)}\right)x_1 \\
&= h(t) x_t - g(t) x_0 + f(t) x_1.
\end{align*}

The model is trained to match this target velocity using the loss:
\[
\mathcal{L}_t(\phi) = \mathbb{E}_{(x_0, x_1)} \, \mathbb{E}_{x_t \sim q_t(\cdot \mid x_0, x_1)} \left[ \| \phi(x_t, t) - v_t(x_t) \|^2 \right],
\]
with gradient:
\[
\nabla_\phi \mathcal{L}(\phi) = \mathbb{E}_{(x_0, x_1)} \, \mathbb{E}_{x_t \sim q_t(\cdot \mid x_0, x_1)} \left[ 2 (\phi(x_t, t) - v_t(x_t)) \right].
\]

A Rao-Blackwellized gradient can be derived by conditioning on \( x_t \):
\begin{align*}
\nabla_\phi \mathcal{L}(\phi) 
&= \mathbb{E}_{x_t} \left[ 2 \left( \phi(x_t, t) - \mathbb{E}_{(x_0, x_1) \mid x_t} \left[h(t)x_t - g(t)x_0 + f(t)x_1\right] \right) \right] \\
&= \mathbb{E}_{x_t} \left[ 2 \left( \phi(x_t, t) - h(t)x_t + g(t) \mathbb{E}_{x_0 \mid x_t}[x_0] - f(t) \mathbb{E}_{x_1 \mid x_t}[x_1] \right) \right] 
\end{align*}

We use a single sample \(x_0\) to estimate \(\mathbb{E}[x_0 \mid x_t]\) then use the same estimation technique as in Orbit Diffusion to compute \(\mathbb{E}[x_1 \mid x_t]\), enabling efficient Rao-Blackwellized gradient estimation.


\subsubsection{MCG - Training Protocol}

Instead of training from scratch, we finetune ETFlow using their public checkpoint. During training, we explicitly incorporate both forms of symmetry relevant to molecular data: discrete graph automorphisms and continuous spatial rotations. For each molecule, we uniformly sample \(50\) elements from its automorphism group using the \texttt{pynauty} library, capturing permutation symmetries in the 2D molecular graph structure. Simultaneously, we sample \(200\) elements from the rotation group \(\mathsf{SO}(3)\) to account for the continuous rotational symmetries of its 3D conformation. This symmetry-aware augmentation is applied consistently across the dataset to ensure that the model learns to respect and exploit both types of equivariances. All other training settings, including optimizer configurations and learning rate schedules, follow the defaults of \textsc{ETFlow}.

\subsubsection{MCG - Evaluation Protocol}

\label{sub_app:QM9_metrics}

In the test set, for each molecule with \(L\) ground-truth conformers, we generate \(K = 2L\) conformers and evaluate their quality using standard metrics.

\paragraph{Evaluation Metrics.} As a conformer $C$ represents an assignment of each atom in the molecular graph to a point in 3D space, it can be viewed as a set of vectors in $\mathbb{R}^{3n}$.To evaluate molecular conformer generation, previous works have employed two key metrics: Average Minimum RMSD (AMR) and Coverage (COV) for both Precision (P) and Recall (R). Given a molecular graph, we generate twice as many conformers as those provided by CREST. Let:
\begin{itemize}
    \item $\{C^*_l\}^L_{l=1}$ be the set of grounth-truth conformers provided by CREST.
    \item $\{C^*_k\}^K_{k=1}$ be the set of generated conformers, where $K=2L$.
    \item $\delta$ be a predefined RMSD threshold for considering a conformer match.
\end{itemize}

\textbf{COV-P}: Measures the proportion of generated conformers that closely match at least one ground-truth conformer.
\[
\text{COV-P} = \frac{1}{K} \left| \{ k \in [1, K] \mid \exists l \in [1, L], \text{RMSD}(C_k, C_l^*) < \delta \} \right|
\]

\textbf{AMR-P}: Computes the average of the minimum RMSD values between each generated conformer and its closest ground-truth conformer.
\[
\text{AMR-P} = \frac{1}{K} \sum_{k=1}^{K} \min_{l=1}^{L} \text{RMSD}(C_k, C_l^*)
\]

\textbf{COV-R}: Measures the proportion of ground-truth conformers that have at least one close-enough generated conformer.
\[
\text{COV-R} = \frac{1}{L} \left| \{ l \in [1, L] \mid \exists k \in [1, K], \text{RMSD}(C_k, C_l^*) < \delta \} \right|
\]

\textbf{AMR-R}: Computes the average of the minimum RMSD values between each ground-truth conformer and its closest generated conformer.
\[
\text{AMR-R} = \frac{1}{L} \sum_{k=1}^{L} \min_{l=1}^{K} \text{RMSD}(C_k, C_l^*)
\]

\subsubsection{MCG - Full results}
\label{sub_app:QM9_full_results}
\begin{figure}
    \centering
        \includegraphics[width=0.5\linewidth]{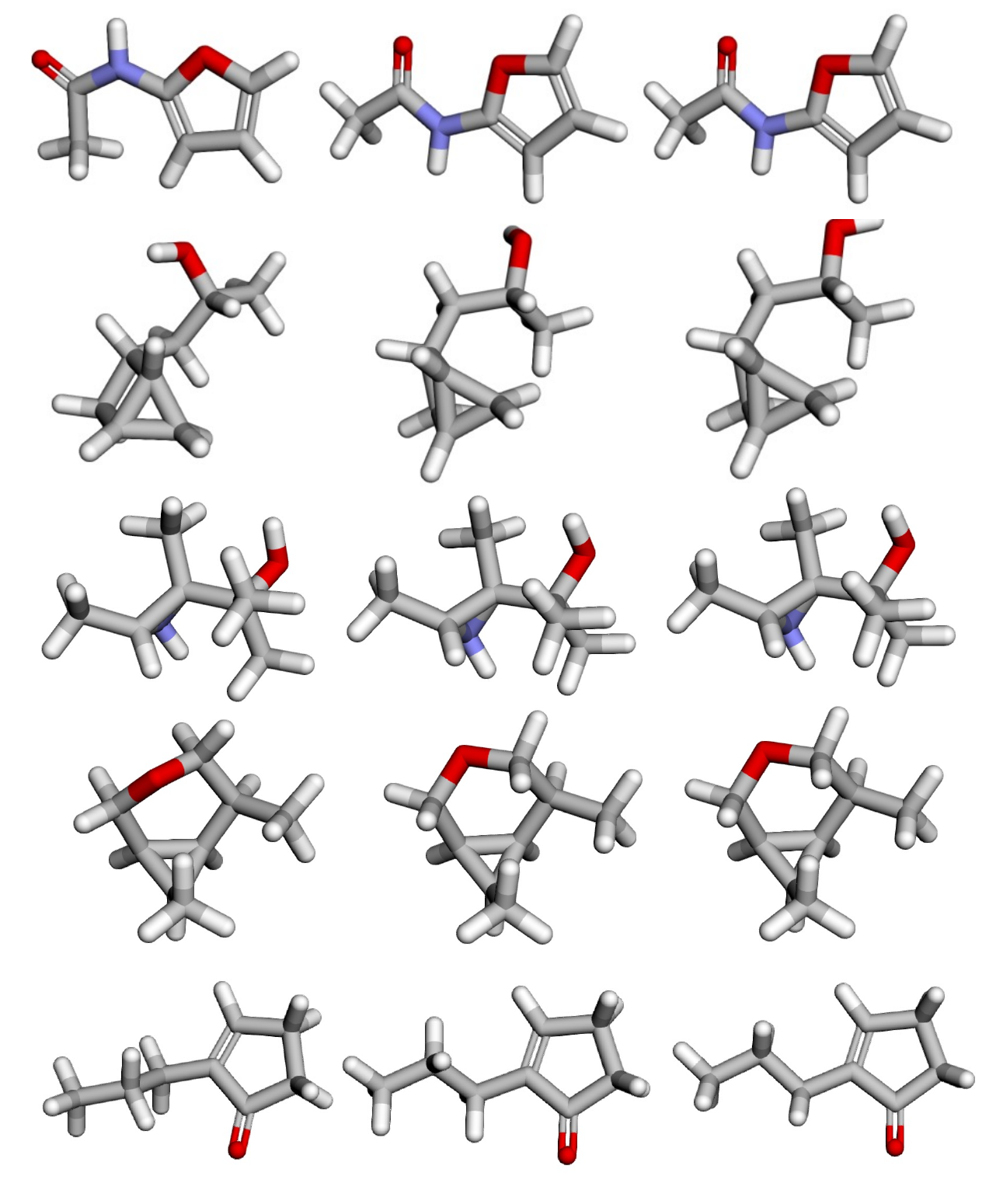}
        \caption{Molecular conformers generated by \textsc{ETFlow} (left), + [\texttt{OrbDiff}] (center), and ground-truth (right).}
        \label{fig:qm9_viz_appendix}
\end{figure}

\input{DELTA_sections/Experiments/table_qm9}

\subsection{Crystal Structure Prediction (CSP)}
\label{sub_app:CSP}

Crystal Structure Prediction (CSP) is the process of identifying the most stable three-dimensional arrangement of atoms in a crystalline solid, given only its chemical formula. This task lies at the heart of computational materials science, as the resulting crystal structure dictates key physical and chemical properties—including thermodynamic stability, electronic conductivity, and chemical reactivity~\citepcolor{Wei2024CSPBenchABA, Shao2021ASDA, kim2024mofflow}. Despite its importance, CSP remains a formidable challenge due to the immense combinatorial search space of atomic positions and the presence of complex symmetry constraints that define equivalent configurations~\citepcolor{oganov2006crystal}. Efficiently navigating this landscape to discover low-energy, physically plausible structures continues to be a central focus of the field.

We also evaluate our method on a related task introduced by TGDMat~\citepcolor{das2025periodic}, where crystal structures are generated based on textual descriptions of the desired materials. This setting reflects a more user-centric interface for materials design, where scientists or domain experts can specify target properties or structural features in natural language. The task includes two levels of textual conditioning: long descriptions, which provide detailed structural and compositional information, and short descriptions, which are more concise and easier to obtain. Supporting text-to-structure generation enables more accessible and flexible workflows in materials discovery, especially in scenarios where precise structural data may be unavailable. 

\subsubsection{CSP - Dataset}

We evaluate our method on two used CSP benchmarks: Perov-5~\citepcolor{castelli2012new, castelli2012computational} and MP-20~\citepcolor{jain2013commentary}. These datasets encompass a broad range of inorganic crystal compositions. Perov-5 comprises 18,928 perovskite structures, which share a common structural motif but vary in elemental composition. In contrast, MP-20 includes 45,231 stable inorganic materials curated from the Materials Project database~\citepcolor{jainmaterials}, offering a more diverse set of crystal systems and chemistries.

\subsubsection{CSP - Baselines}


For comparison, we use the three strongest baselines from the DiffCSP paper~\citepcolor{jiao2023crystal}: P-cG-SchNet~\citepcolor{diffcsp_bl1}, CDVAE~\citepcolor{diffcsp_bl2}, and DiffCSP, all with publicly available implementations. 


\subsubsection{CSP — DiffCSP and TGDMat with Orbit Diffusion}
\label{sub_app:csp_wn}

DiffCSP~\citepcolor{jiao2023crystal} is a diffusion-based framework for crystal structure prediction that jointly models lattice parameters and atomic positions while respecting the fundamental symmetries of crystalline materials. Since TGDMat is built upon DiffCSP, we focus on describing how Orbit Diffusion integrates with DiffCSP; the same integration applies directly to TGDMat.

DiffCSP formulates the task as a \textit{joint diffusion process} with two interconnected components: one for the \textit{lattice} and one for the \textit{atomic coordinates}. The lattice defines the shape and scale of the unit cell, while the coordinates specify atomic positions in fractional units relative to the lattice vectors. To capture the relevant symmetries, the lattice diffusion is \(\mathsf{O}(3)\)-equivariant (invariant under rotations and reflections), and the coordinate diffusion is both \textit{permutation equivariant} and \textit{periodic translation equivariant}, reflecting atomic indistinguishability and lattice periodicity.

Since the lattice involves only a few parameters, it is a relatively simple subtask. We therefore concentrate on the more challenging component: generating atomic coordinates. To ensure periodic translation equivariance, DiffCSP defines a forward diffusion process based on the kernel \(\hat{q}^G(x_t \mid x_0)\), modeled as a \textit{Wrapped Normal} distribution—a periodic analogue of the Gaussian—ensuring the diffusion respects the toroidal geometry of fractional coordinates. Specifically,
\begin{align}
    \hat{q}^G(x_t \mid x_0) \propto \sum_{z \in \mathbb{Z}^d} \exp \left( -\frac{\|x_t - x_0 + z\|^2}{2\sigma_t^2} \right),
\end{align}
which defines a valid density on the torus \(\mathbb{T}^d\). One can verify that \(\hat{q}^G(g \circ x_t \mid g \circ x_0) = \hat{q}^G(x_t \mid x_0)\) for any \(g\) in the periodic translation group~\citepcolor{jiao2023crystal}, confirming its equivariance. Consequently, all our theoretical results (e.g., \cref{the:theo_1} and \cref{theo_2}) hold when applying Orbit Diffusion to DiffCSP under the periodic translation group.

\paragraph{Orbit Diffusion with non-uniform group sampling.}

Rather than sampling uniformly from the group \(G\), for the periodic translation group we propose sampling translation elements from a Wrapped Normal distribution. We refer to this variant as \texttt{OrbDiff\_WN}. Formally,
\begin{align}
    \nu_t(g) \propto \sum_{z_g \in \mathbb{Z}^{d_g}} \exp \left( -\frac{\| m_g + z_g \|^2}{2 \sigma_g(t)^2 } \right),
\end{align}
where \(m_g\) is the translation vector corresponding to group element \(g\), and \(\sigma_g(t)\) is the time-dependent bandwidth. In our experiments with both DiffCSP and TGDMat, we set \(\sigma_g(t) = 2\sigma_t\). To sample from this distribution, we first draw \(\epsilon \sim \mathcal{N}(0, I)\) in \(\mathbb{R}^3\), and then compute \(m_g = \sigma_g(t) \epsilon \mod 1\). We sample 1000 such group elements per step to form the group approximation. 


\paragraph{More training details.}
All models were trained on a single NVIDIA GeForce RTX 4090 GPU. TGDMat was trained for 1,500 epochs, while DiffCSP was trained for 500 epochs. On the MP20 dataset, each epoch took roughly 15 seconds, resulting in total training times of 6.25 hours for TGDMat and 2.08 hours for DiffCSP. On the Perov-5 dataset, each epoch took about 5 seconds, corresponding to 2.08 hours (TGDMat) and 0.69 hours (DiffCSP) of training time.

\subsubsection{CSP - Evaluation Protocol}
\label{sub_app:CSP_metrics}

To evaluate crystal structure prediction, we randomly generate one sample for each structure in the test set. We then calculate two metrics: the Match Rate and the average Root Mean Square Distance (RMSD) across the test set. We repeat this procedure three times and report the median values of these metrics for more reliability. 

\textbf{Match rate}: The Match Rate is defined as the proportion of predicted structures that successfully match the corresponding ground-truth structures in the test set. Specifically, it is calculated as follows:

\begin{equation*}
\text{Match Rate} = \frac{\text{Number of matched structure pairs}}{\text{Total number of test samples}}
\end{equation*}

Following previous works ~\citepcolor{jiao2023crystal, diffcsp_bl2}, we use the \texttt{StructureMatcher} class from the \texttt{pymatgen} library to determine structure matching. The matching process is based on the following criteria:
\begin{itemize}
    \item Length Tolerance (\textit{ltol}): 0.5 (fractional length tolerance).
    \item Site Tolerance (\textit{stol}): 0.3 (fraction of the average free length per atom)
    \item Angle Tolerance (\textit{atol}): 10 (in degrees)
\end{itemize}
The \texttt{StructureMatcher} algorithm aligns the lattice vectors of two structures. If the tolerance criteria are satisfied, the structures are considered matched.

\textbf{RMSD}: For an alignment between lattices of two structures, \texttt{StructureMatcher} continues to align atoms to compute the average RMSD. The process is repeated for all possible lattices to find the smallest RMSD. Then the Average RMSD is computed as the average of the smallest RMSD of all matched structure pairs. 
\begin{equation*}
\text{RMSD} = \frac{1}{N_{\text{matched}}} \sum_{i=1}^{N_{\text{matched}}} \text{RMSD}(\text{generated}_i, \text{ground-truth}_i)
\end{equation*}

Here, $N_{\text{matched}}$ is the total number of matched structures. Unmatched structures are excluded from the calculation.

Ideally, we aim for a high Match Rate and a low RMSD. A low Match Rate with a low RMSD is not useful because unmatched samples are effectively treated as having very high RMSD. Thus, RMSD alone cannot fully capture prediction quality. We emphasize that Match Rate is more critical, especially during initial screening, where we prioritize valid structures over perfectly matched ones.

\subsubsection{CSP - Full Results and Visualizations}

\begin{figure}[H]
    \centering
    \includegraphics[]{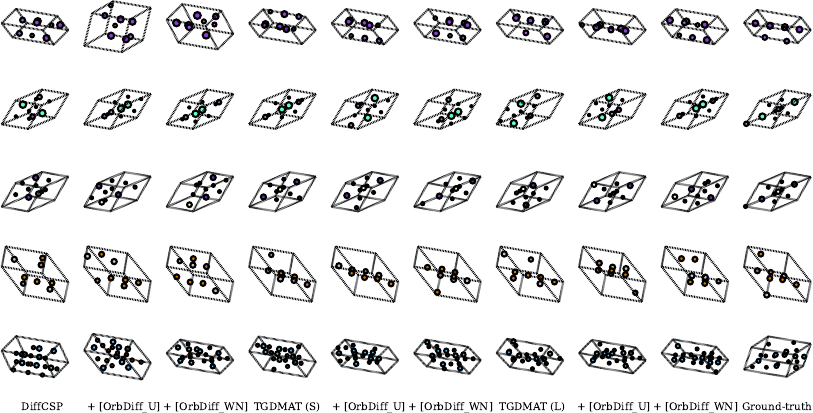}
    \caption{Qualitative comparison of Crystal Structure Predictions by 9 models, including DiffCSP, TGDMAT short and TGDMAT long with baselines,  \texttt{OrbDiff\_U}, and \texttt{OrbDiff\_WN} against ground-truth samples on randomly selected samples from MP-20 dataset.}
    \label{fig:viz_mp20}
\end{figure}

\begin{figure}[H]
    \centering
    \includegraphics[]{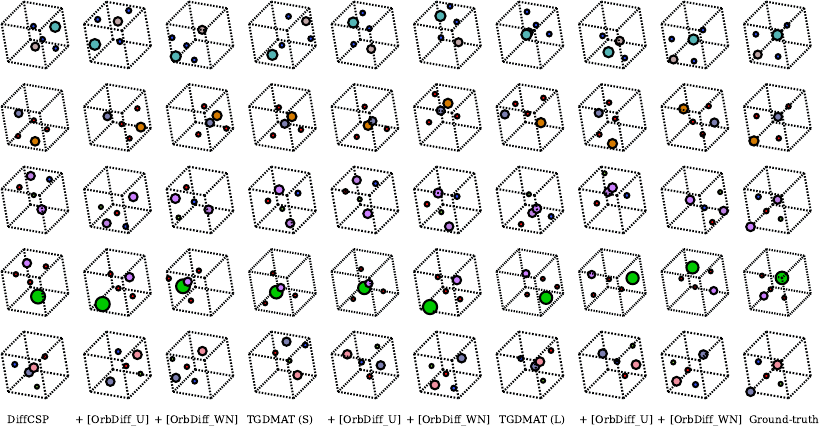}
    \caption{Qualitative comparison of Crystal Structure Predictions by 9 models, including DiffCSP, TGDMat (S) and TGDmat (L) with baselines, \texttt{OrbDiff\_U}, and \texttt{OrbDiff\_WN} against ground-truth samples on randomly selected samples from Perov-5 dataset.}
    \label{fig:viz_perov5}
\end{figure}

For the TGDMat models (TGDMat (S) and TGDMat (L)), we trained both from scratch, experimenting with and without the proposed loss functions: \texttt{[OrbDiff\_U]} and \texttt{[OrbDiff\_WN]}. Meanwhile, the DiffCSP model was trained using our proposed losses, while the baseline models (P-cG-SchNet, CDVAE, and DiffCSP) rely on the results reported in the original DiffCSP paper. Quantitative results are summarized in Tables ~\ref{tab:tgdmat_appendix} and ~\ref{tab:diffcsp_appendix} with qualitative comparisons shown in Figures~\ref{fig:viz_mp20} and \ref{fig:viz_perov5}.

\input{DELTA_sections/Experiments/table_diffcsp_appendix}

\subsection{Protein Structure Generation (PSG) with Non-Equivariant Denoiser \textsc{ProteiNA}}

Protein structure generation focuses on sampling physically valid 3D conformations of proteins by learning a probabilistic distribution over either atomistic or coarse-grained representations. This generative task plays a crucial role in de novo protein design and has broad implications for understanding protein folding and function~\citepcolor{PaoHuang2023EigenFoldGPA, Wu2022ProteinSGA, Watson2023DeNDA, Huguet2024SequenceAugmentedSMA, PaoHuang2023EigenFoldGPA}. Unlike traditional structure prediction, which aims to infer a single or most likely conformation, generative models must capture the full structural manifold—accounting for the inherent $\mathsf{SO}(3)$ rotational symmetry of protein backbones, maintaining biochemical realism, and respecting physical constraints such as bond lengths and steric clashes~\citepcolor{Gaujac2024LearningTLA, proteina}. Effectively modeling these aspects ensures that generated structures are not only diverse but also biologically and physically plausible.

\subsubsection{PSG - Baselines}

We list the baselines used by the work of \citepcolor{proteina} as follows:

\begin{itemize}
  \item FrameDiff~\citepcolor{yim2023se}
  \item FoldFlow~\citepcolor{bose2023se}
  \item FrameFlow~\citepcolor{yim2023fast}
  \item ESM3~\citepcolor{hayes2024simulating}
  \item Chroma~\citepcolor{Chroma2023}
  \item RFDiffusion~\citepcolor{Watson2023DeNDA}
  \item Proteus~\citepcolor{wang2024proteus}
  \item Genie2~\citepcolor{lin2024genie2}
\end{itemize}

\paragraph{\textsc{ProteiNA}.}  
\textsc{ProteiNA} is a large-scale, flow-based model for generating protein backbones, built on a scalable transformer architecture. Unlike equivariant models, it does not enforce equivariance, granting more architectural flexibility. This choice allows the use of powerful transformer networks with hundreds of millions of parameters, enabling \textsc{ProteiNA} to effectively learn from large datasets. As a result, it excels at modeling complex protein structures by balancing expressiveness and computational scalability.

There are three variants of \textsc{ProteiNA}:\\
(i)~\(\mathcal{M}_\text{FS}\), a 200M-parameter transformer with an additional 15M parameters in triangle layers, trained on the \textbf{Foldseek AFDB clusters} dataset \(\mathcal{D}_\text{FS}\), which includes 555{,}318 structures of lengths 32--256;\\
(ii)~\(\mathcal{M}_\text{FS}^\text{no-tri}\), a simplified version without triangle layers, also with 200M parameters and trained on the same dataset;\\
(iii)~\(\mathcal{M}_\text{21M}\), a 400M-parameter transformer with 15M triangle parameters, trained on a \textbf{high-quality filtered AFDB subset} \(\mathcal{D}_{21M}\), comprising approximately 21M structures. \(\mathcal{M}_\text{21M}\) represents the current state of the art in designability modeling.
\vspace{-0.2cm}

\subsubsection{\textsc{ProteiNA} with Orbit Diffusion}
We apply Orbit Diffusion to the simplest variant of \textsc{ProteiNA}, namely \(\mathcal{M}_\text{FS}^\text{no-tri}\), by fine-tuning the public checkpoint released by~\citetcolor{proteina}, since we do not have access to the extensive compute resources used for the original training. Our fine-tuning setup uses 4 A100 GPUs for 24 hours on the same dataset, whereas the original training employed 96 GPUs. 

The key symmetry group in protein structure generation is \(\mathsf{SO}(3)\), which represents 3D rotations. To exploit this symmetry using \texttt{OrbDiff}, we apply 10,000 uniformly sampled random rotations to each training sample as part of our Rao-Blackwell estimator, enhancing both the efficiency and stability of the flow matching process.

The only change we introduce to the original model is replacing the conditional flow matching loss with our proposed flow matching objective (see~\cref{sub_app:orb_flow}). 

For a fair comparison, we also fine-tune \(\mathcal{M}_\text{FS}^\text{no-tri}\) (+ [finetune]) using the original loss under the same computational budget, with a batch size of 8 and 32 gradient accumulation steps.

\vspace{-0.2cm}
\subsubsection{PSG - Evaluation Protocol}
\label{sub_app:PSG_eval}

To evaluate the quality of our generated protein backbones, we rely on three widely used metrics: designability, diversity, and novelty. Following the protocol established by~\citetcolor{proteina}, we generate 500 samples total—100 for each length in \{50, 100, 150, 200, 250\}—and compute all metrics on this dataset. Among these, \textbf{designability is the most critical metric}, as it directly reflects the biological feasibility of the generated structures and serves as the foundation for the other two metrics.

\vspace{-0.2cm}
\paragraph{Designability.}  
Designability measures whether a backbone structure can realistically be encoded by an amino acid sequence. For each generated backbone, we produce 8 candidate sequences using ProteinMPNN~\citepcolor{dauparas2022robust} with a sampling temperature of 0.1. These sequences are then folded using ESMFold~\citepcolor{lin2023evolutionary}, and the root mean square deviation (RMSD) is calculated between each predicted structure and the original backbone. A backbone is deemed designable if at least one sequence folds with an RMSD below 2\AA, where this minimum RMSD is known as the self-consistency RMSD (scRMSD).

Since \textbf{diversity and novelty are computed only on designable samples}, accurate assessment of designability is essential for interpreting the other metrics meaningfully.

The designability score for a model is reported as the fraction of samples deemed designable. Additionally, we report the average scRMSD across all samples, allowing for a more nuanced comparison between our model and existing baselines.

\paragraph{Diversity.}  
We evaluate diversity among the designable samples using two approaches. First, we compute the average pairwise TM-score within each length group (50, 100, 150, 200, and 250) as a measure of structural variation. Lower average TM-scores indicate greater diversity. 

Second, we calculate diversity (cluster) by grouping designable samples into clusters based on a TM-score threshold of 0.5. Each cluster contains samples with pairwise TM-scores above this threshold. Diversity (cluster) is then defined as the ratio of the total number of clusters to the number of designable samples. A higher ratio reflects a larger number of distinct structural groups relative to the sample size, signaling increased diversity.

\paragraph{Novelty.}  
Novelty measures how structurally distinct the designable samples are compared to known protein structures. For each designable backbone, we compute the TM-score to its closest match in two reference sets: the Protein Data Bank (PDB) and the \(\mathcal{D}_\text{FS}\) dataset used for training. The average of these maximum TM-scores across all designable samples is reported as the novelty score. Lower values indicate the model generates structures that are more novel relative to both established experimental data and the training distribution.

\subsubsection{PSG - Full results}
\label{sub_app:PSG_full}

\vspace{-0.5cm}

\begin{table}[ht]
\centering
\caption{\textbf{Protein Structure Generation.} Full comparison with baseline models; all baseline results are taken from \citepcolor{proteina}. ``+ [finetune]'' indicates $\mathcal{M}_{FS}^{\text{no-tri}}$ finetuned with the original loss, while ``+ [\texttt{OrbDiff}]'' denotes finetuning with \texttt{OrbDiff}.}

\resizebox{0.9\textwidth}{!}{
\begin{tabular}{lccccccc}
\toprule
\multirow{2}{*}{Model} &  
\multicolumn{2}{>{\columncolor[HTML]{f2cc8f}}c}{Designability} & 
\multicolumn{2}{c}{Diversity} & 
\multicolumn{2}{c}{Novelty} \\
\cmidrule(lr){2-3} \cmidrule(lr){4-5} \cmidrule(lr){6-7}
& Fraction~(↑) & scRMSD~(↓) & Cluster~(↑) & TM-score~(↓) & PDB~(↓) & AFDB~(↓) \\
\midrule
FrameDiff & 65.4 & - & 0.39 & 0.40 & 0.73 & 0.75 \\
FoldFlow (base) & 96.6 & - & 0.42 & 0.75 & 0.75 & 0.77 \\
FoldFlow (stoc.) & 97.0 & - & 0.61 & 0.38 & 0.62 & 0.68 \\
FoldFlow (OT) & 97.2 & - & 0.37 & 0.41 & 0.71 & 0.75 \\
FrameFlow & 88.6 & - & 0.59 & 0.34 & 0.79 & 0.80 \\
ESM3 & 22.0 & - & 0.52 & 0.57 & 0.70 & 0.75 \\
Chroma & 78.8 & - & 0.42 & 0.43 & 0.77 & 0.76 \\
RFDiffusion & 94.4 & - & 0.46 & 0.34 & 0.79 & 0.80 \\
Proteus & 94.4 & - & 0.42 & 0.43 & 0.77 & 0.80 \\
Genie2 & 95.2 & - & 0.59 & 0.38 & \textbf{0.63} & 0.69 \\
\(\mathcal{M}_\text{21M}\) & \textbf{99.0} & \textbf{0.72} & 0.30 & 0.39 & 0.81 & 0.84 \\
\midrule
\(\mathcal{M}^\text{no-tri}_\text{FS}\) & 93.8 & 1.04 & \textbf{0.62} & \textbf{0.36} & \textbf{0.69} & \textbf{0.76} \\
+ [finetune] & 93.8 & 1.00 & 0.54 & 0.37 & 0.74 & 0.83 \\
\rowcolor[HTML]{EAEAEA}
+ [\texttt{OrbDiff}] & \textbf{95.6} & \textbf{0.93} & 0.52 & 0.37 & 0.74 & 0.83 \\
\bottomrule
\end{tabular}
}
\label{tab:model_comparison}
\end{table}





The state-of-the-art model $\mathcal{M}_\text{21M}$, with 400M parameters and trained on a large, high-quality dataset, achieves the highest designability (99.0\%) and lowest scRMSD (0.72), reflecting its strong reconstruction capability. However, this comes at the expense of diversity and novelty: it exhibits the lowest diversity score (Cluster: 0.30) and higher novelty metrics (PDB: 0.81, AFDB: 0.84), indicating reduced structural variety and generalization.

In comparison, our base model $\mathcal{M}^\text{no-tri}_\text{FS}$ already achieves competitive performance (designability: 93.8\%, Cluster: 0.62, PDB: 0.69), and naive finetuning with data augmentation (``+ [finetune]'') fails to improve designability or diversity. Notably, our method (``+ [\texttt{OrbDiff}]'') improves designability to 95.6\% and reduces scRMSD to 0.93, while preserving competitive diversity (Cluster: 0.52) and novelty (PDB: 0.74). This highlights that \texttt{OrbDiff} is an effective finetuning strategy to enhance functional accuracy without fully sacrificing structural diversity—achieving a better trade-off than both naive finetuning and heavily overparameterized models.

%% file: DELTA_sections/Experiments/table_qm9.tex
\begin{table}[H]
    \caption{Molecular conformer generation performance on GEOM-QM9.  
    \footnotesize  
    \textsuperscript{*} Reported in the original paper.  
    \textsuperscript{†} Obtained using the published checkpoint.  
    \textsuperscript{‡} From our reimplementation trained from scratch.  
    }
    \label{tab:qm9}
    \centering
    \setlength{\tabcolsep}{1.4pt}
    \renewcommand{\arraystretch}{0.9}
    \resizebox{\textwidth}{!}{
    \begin{tabular}{lcccccccccccc}
    \toprule
    \multirow{2}{*}{Models} & \multicolumn{6}{c}{Recall} & \multicolumn{6}{c}{Precision} \\
    \cmidrule(lr){2-7} \cmidrule(lr){8-13}
     & \multicolumn{2}{c}{Cov@0.1 (↑)} & \multicolumn{2}{c}{Cov@0.5 (↑)} & \multicolumn{2}{c}{AMR (↓)} 
     & \multicolumn{2}{c}{Cov@0.1 (↑)} & \multicolumn{2}{c}{Cov@0.5 (↑)} & \multicolumn{2}{c}{AMR (↓)} \\
    \cmidrule(lr){2-3} \cmidrule(lr){4-5} \cmidrule(lr){6-7} \cmidrule(lr){8-9} \cmidrule(lr){10-11} \cmidrule(lr){12-13}
     & Mean & Median & Mean & Median & Mean & Median & Mean & Median & Mean & Median & Mean & Median \\
    \midrule
    \textsc{GeoDiff } & - & - & 76.5 & \textbf{100.0} & 0.297 & 0.229 & - & - & 50.0 & 33.5 & 1.524 & 0.510 \\
    \textsc{GeoMol\textsuperscript{†} } & 28.4 & 0.0 & 91.1 & \textbf{100.0} & 0.224 & 0.194 & 20.7 & 0.0 & 85.8 & \textbf{100.0} & 0.271 & 0.243 \\
    Torsional Diff.\textsuperscript{†}  & 37.7 & 25.0 & 88.4 & \textbf{100.0} & 0.178 & 0.147 & 27.6 & 12.5 & 84.5 & \textbf{100.0} & 0.221 & 0.195 \\
    \textsc{MCF\textsuperscript{†} } & 81.9 & \textbf{100.0} & 94.9 & \textbf{100.0} & 0.103 & 0.049 & 78.6 & \textbf{93.8} & \textbf{93.9} & \textbf{100.0} & \textbf{0.113} & 0.055 \\
    \midrule
    \textsc{ETFlow\textsuperscript{*}} & - & - & 96.5 & \textbf{100.0} & 0.073 & 0.047 & - & - & 94.1 & \textbf{100.0} & 0.098 & 0.039 \\
    \textsc{ETFlow\textsuperscript{†}  } & 79.5 & \textbf{100.0} & 93.8 & \textbf{100.0} & 0.096 & 0.037 & 74.4 & 83.3 & 88.7 & \textbf{100.0} & 0.142 & 0.066 \\
    \textsc{ETFlow\textsuperscript{‡} } & 81.4 & \textbf{100.0} & 94.4 & \textbf{100.0} & 0.092 & 0.039 & 74.6 & 85.5 & 89.1 & \textbf{100.0} & 0.145 & 0.064 \\

    \rowcolor[HTML]{EAEAEA}
    + {[\texttt{OrbDiff}]} & \textbf{{85.4}} & \textbf{100.0} & \textbf{{96.3}} & \textbf{100.0} & \textbf{{0.074}} & \textbf{0.027} & \textbf{{80.2}} & \textbf{{93.9}} & {91.9} & \textbf{100.0} & \textbf{{0.113}} & \textbf{0.042} \\
    \bottomrule
    \end{tabular}}
\end{table}

%% file: DELTA_sections/Experiments/table_diffcsp_appendix.tex
\begin{table}[H]
\centering
\begin{minipage}{0.48\textwidth}
\centering
\caption{Text-guided CSP with TGDMat.}
\label{tab:tgdmat_appendix}
\resizebox{\linewidth}{!}{ 
\begin{tabular}{lcccc}
\toprule
\multirow{2}{*}{Method} & \multicolumn{2}{c}{Perov-5} & \multicolumn{2}{c}{MP-20} \\
\cmidrule(lr){2-3} \cmidrule(lr){4-5}
& Match (↑) & RMSE (↓) & Match (↑) & RMSE (↓) \\
\midrule
TGDMat (S)         &  59.39 & 0.066 & 59.90 & 0.078  \\
\rowcolor[HTML]{EAEAEA}
+ {[\texttt{OrbDiff\_U}]}   &   63.51 &  0.062 & 56.50 & 0.085 \\
\rowcolor[HTML]{EAEAEA}
+ {[\texttt{OrbDiff\_WN}]}  &\textbf{65.57}  & \textbf{0.054} &  \textbf{61.29} & \textbf{0.072} \\
\midrule
TGDMat (L)          & 95.17 & 0.013 &  61.91 &  0.081 \\
\rowcolor[HTML]{EAEAEA}
+ {[\texttt{OrbDiff\_U}]}   &  95.88  & \textbf{0.012} & 65.94 & \textbf{0.069} \\
\rowcolor[HTML]{EAEAEA}
+ {[\texttt{OrbDiff\_WN}]}  & \textbf{95.98} & \textbf{0.012} & \textbf{66.74} & \textbf{0.069}  \\
\bottomrule
\end{tabular}
}
\end{minipage}
\hfill
\begin{minipage}{0.48\textwidth}
\centering
\caption{Crystal Structure Prediction (CSP). }
\label{tab:diffcsp_appendix}
\resizebox{\linewidth}{!}{ 
\begin{tabular}{lcccc}
\toprule
\multirow{2}{*}{Method} & \multicolumn{2}{c}{Perov-5} & \multicolumn{2}{c}{MP-20} \\
\cmidrule(lr){2-3} \cmidrule(lr){4-5}
& Match (↑) & RMSE (↓) & Match (↑) & RMSE (↓) \\
\midrule
P-cG-SchNet     & 48.22 & 0.418  & 15.39 & 0.376 \\
CDVAE           & 45.31 & 0.114  & 33.90 & 0.105  \\
\midrule
DiffCSP         & 52.02 & 0.076  & 51.49 & 0.063 \\
\rowcolor[HTML]{EAEAEA}
+ {[\texttt{OrbDiff\_U}]}  & 52.29&0.078  & 54.47 & 0.054 \\
\rowcolor[HTML]{EAEAEA}
+ {[\texttt{OrbDiff\_WN}]} & \textbf{52.39} & \textbf{0.069} & \textbf{55.70} & \textbf{0.053}  \\
\bottomrule
\end{tabular}
}
\end{minipage}
\end{table}

%% file: neurips_2025.bbl
\begin{thebibliography}{74}
\providecommand{\natexlab}[1]{#1}
\providecommand{\url}[1]{\texttt{#1}}
\expandafter\ifx\csname urlstyle\endcsname\relax
  \providecommand{\doi}[1]{doi: #1}\else
  \providecommand{\doi}{doi: \begingroup \urlstyle{rm}\Url}\fi

\bibitem[Abbe \& Boix-Adser{\`{a}}(2022)Abbe and Boix-Adser{\`{a}}]{nonuniversalityequi}
Emmanuel Abbe and Enric Boix-Adser{\`{a}}.
\newblock On the non-universality of deep learning: Quantifying the cost of symmetry.
\newblock In \emph{Advances in Neural Information Processing Systems 35 (NeurIPS 2022)}, 2022.

\bibitem[Abramson et~al.(2024{\natexlab{a}})Abramson, Adler, Dunger, Evans, Green, Pritzel, Ronneberger, Willmore, Ballard, Bambrick, et~al.]{abramson2024accurate}
Josh Abramson, Jonas Adler, Jack Dunger, Richard Evans, Tim Green, Alexander Pritzel, Olaf Ronneberger, Lindsay Willmore, Andrew~J Ballard, Joshua Bambrick, et~al.
\newblock Accurate structure prediction of biomolecular interactions with alphafold 3.
\newblock \emph{Nature}, 2024{\natexlab{a}}.

\bibitem[Abramson et~al.(2024{\natexlab{b}})Abramson, Adler, Dunger, Evans, Green, Pritzel, Ronneberger, Willmore, Ballard, Bambrick, et~al.]{alphafold3}
Josh Abramson, Jonas Adler, Jack Dunger, Richard Evans, Tim Green, Alexander Pritzel, Olaf Ronneberger, Lindsay Willmore, Andrew~J Ballard, Joshua Bambrick, et~al.
\newblock Accurate structure prediction of biomolecular interactions with alphafold 3.
\newblock \emph{Nature}, 630\penalty0 (8016):\penalty0 493--500, 2024{\natexlab{b}}.

\bibitem[Anand \& Achim(2022)Anand and Achim]{protein_diff}
Namrata Anand and Tudor Achim.
\newblock Protein structure and sequence generation with equivariant denoising diffusion probabilistic models.
\newblock \emph{CoRR}, abs/2205.15019, 2022.

\bibitem[Axelrod \& G{\'o}mez-Bombarelli(2020)Axelrod and G{\'o}mez-Bombarelli]{Axelrod2020MolecularMLA}
Simon Axelrod and Rafael G{\'o}mez-Bombarelli.
\newblock Molecular machine learning with conformer ensembles.
\newblock \emph{Machine Learning: Science and Technology}, 4, 2020.
\newblock URL \url{https://api.semanticscholar.org/CorpusId:229181029}.

\bibitem[Axelrod \& Gomez-Bombarelli(2022)Axelrod and Gomez-Bombarelli]{axelrod2022geom}
Simon Axelrod and Rafael Gomez-Bombarelli.
\newblock Geom, energy-annotated molecular conformations for property prediction and molecular generation.
\newblock \emph{Scientific Data}, 9\penalty0 (1):\penalty0 185, 2022.

\bibitem[Batatia et~al.(2022)Batatia, Kovacs, Simm, Ortner, and Cs{\'a}nyi]{batatia2022mace}
Ilyes Batatia, David~P Kovacs, Gregor Simm, Christoph Ortner, and G{\'a}bor Cs{\'a}nyi.
\newblock Mace: Higher order equivariant message passing neural networks for fast and accurate force fields.
\newblock \emph{Advances in neural information processing systems}, 2022.

\bibitem[Bose et~al.(2024)Bose, Akhound-Sadegh, Huguet, Fatras, Rector-Brooks, Liu, Nica, Korablyov, Bronstein, and Tong]{bose2023se}
Avishek~Joey Bose, Tara Akhound-Sadegh, Guillaume Huguet, Kilian Fatras, Jarrid Rector-Brooks, Cheng-Hao Liu, Andrei~Cristian Nica, Maksym Korablyov, Michael Bronstein, and Alexander Tong.
\newblock Se(3)-stochastic flow matching for protein backbone generation.
\newblock In \emph{Proceedings of the 12th International Conference on Learning Representations (ICLR)}, 2024.

\bibitem[Brehmer et~al.(2024)Brehmer, Behrends, de~Haan, and Cohen]{doesequimatteratscale}
Johann Brehmer, Sönke Behrends, Pim de~Haan, and Taco Cohen.
\newblock Does equivariance matter at scale?
\newblock \emph{arXiv preprint arXiv:2410.23179}, 2024.

\bibitem[Castelli et~al.(2012{\natexlab{a}})Castelli, Landis, Thygesen, Dahl, Chorkendorff, Jaramillo, and Jacobsen]{castelli2012new}
Ivano~E. Castelli, David~D. Landis, Kristian~S. Thygesen, Søren Dahl, Ib~Chorkendorff, Thomas~F. Jaramillo, and Karsten~W. Jacobsen.
\newblock New cubic perovskites for one- and two-photon water splitting using the computational materials repository.
\newblock \emph{Energy \& Environmental Science}, 2012{\natexlab{a}}.

\bibitem[Castelli et~al.(2012{\natexlab{b}})Castelli, Olsen, Datta, Landis, Dahl, Thygesen, and Jacobsen]{castelli2012computational}
Ivano~E Castelli, Thomas Olsen, Soumendu Datta, David~D Landis, S{\o}ren Dahl, Kristian~S Thygesen, and Karsten~W Jacobsen.
\newblock Computational screening of perovskite metal oxides for optimal solar light capture.
\newblock \emph{Energy \& Environmental Science}, 5\penalty0 (2):\penalty0 5814--5819, 2012{\natexlab{b}}.

\bibitem[Chen et~al.(2024)Chen, Katsoulakis, and Zhang]{Theoequi}
Ziyu Chen, Markos~A. Katsoulakis, and Benjamin~J. Zhang.
\newblock Equivariant score-based generative models provably learn distributions with symmetries efficiently.
\newblock \emph{arXiv preprint arXiv:2410.01244}, 2024.

\bibitem[Cohen \& Welling(2016)Cohen and Welling]{equiconv}
Taco Cohen and Max Welling.
\newblock Group equivariant convolutional networks.
\newblock In \emph{Proceedings of the 33rd International Conference on Machine Learning (ICML)}, 2016.

\bibitem[Corso et~al.(2024)Corso, Deng, Polizzi, Barzilay, and Jaakkola]{corso2024discovery}
Gabriele Corso, Arthur Deng, Nicholas Polizzi, Regina Barzilay, and Tommi Jaakkola.
\newblock Deep confident steps to new pockets: Strategies for docking generalization.
\newblock In \emph{International Conference on Learning Representations (ICLR)}, 2024.

\bibitem[DAS et~al.(2025)DAS, Khastagir, Goyal, Lee, Bhattacharjee, and Ganguly]{das2025periodic}
KISHALAY DAS, Subhojyoti Khastagir, Pawan Goyal, Seung-Cheol Lee, Satadeep Bhattacharjee, and Niloy Ganguly.
\newblock Periodic materials generation using text-guided joint diffusion model.
\newblock In \emph{The Thirteenth International Conference on Learning Representations}, 2025.

\bibitem[Dauparas et~al.(2022)Dauparas, Anishchenko, Bennett, Bai, Ragotte, Milles, Wicky, Courbet, de~Haas, Bethel, Leung, Huddy, Pellock, Tischer, Chan, Koepnick, Nguyen, Kang, Sankaran, Bera, King, and Baker]{dauparas2022robust}
Justas Dauparas, Ivan Anishchenko, Nathaniel Bennett, Hua Bai, Robert~J Ragotte, Lukas~F Milles, Basile~IM Wicky, Alexis Courbet, Rob~J de~Haas, Neville Bethel, Patrick~JY Leung, Thomas~F Huddy, Samuel Pellock, Daniel Tischer, Felix Chan, Brian Koepnick, Huy Nguyen, Andrew Kang, Balamurugan Sankaran, Abhishek~K Bera, Neil~P King, and David Baker.
\newblock Robust deep learning--based protein sequence design using proteinmpnn.
\newblock \emph{Science}, 2022.

\bibitem[Deng et~al.(2021)Deng, Litany, Duan, Poulenard, Tagliasacchi, and Guibas]{vectorneuron}
Congyue Deng, Or~Litany, Yueqi Duan, Adrien Poulenard, Andrea Tagliasacchi, and Leonidas~J. Guibas.
\newblock Vector neurons: A general framework for so(3)-equivariant networks.
\newblock In \emph{Proceedings of the 2021 IEEE/CVF International Conference on Computer Vision (ICCV)}, 2021.

\bibitem[Fuchs et~al.(2020)Fuchs, Worrall, Fischer, and Welling]{e3transformer}
Fabian Fuchs, Daniel~E. Worrall, Volker Fischer, and Max Welling.
\newblock Se(3)-transformers: 3d roto-translation equivariant attention networks.
\newblock In \emph{Advances in Neural Information Processing Systems 33 (NeurIPS 2020)}, 2020.

\bibitem[Ganea et~al.(2021)Ganea, Pattanaik, Coley, Barzilay, Jensen, Jr., and Jaakkola]{geomol}
Octavian Ganea, Lagnajit Pattanaik, Connor~W. Coley, Regina Barzilay, Klavs~F. Jensen, William H.~Green Jr., and Tommi~S. Jaakkola.
\newblock Geomol: Torsional geometric generation of molecular 3d conformer ensembles.
\newblock In \emph{Advances in Neural Information Processing Systems 34: Annual Conference on Neural Information Processing Systems 2021, NeurIPS 2021, December 6-14, 2021, virtual}, 2021.

\bibitem[Gaujac et~al.(2024)Gaujac, Dona, Copoiu, Atkinson, Pierrot, and Barrett]{Gaujac2024LearningTLA}
Benoit Gaujac, J'er'emie Dona, Liviu Copoiu, Timothy Atkinson, Thomas Pierrot, and Thomas~D. Barrett.
\newblock Learning the language of protein structure.
\newblock \emph{ArXiv}, abs/2405.15840, 2024.
\newblock URL \url{https://api.semanticscholar.org/CorpusId:270063152}.

\bibitem[Gebauer et~al.(2022)Gebauer, Gastegger, Hessmann, Müller, and Schütt]{diffcsp_bl1}
Niklas W.~A. Gebauer, Michael Gastegger, Stefaan S.~P. Hessmann, Klaus-Robert Müller, and Kristof~T. Schütt.
\newblock Inverse design of 3d molecular structures with conditional generative neural networks.
\newblock \emph{Nature Communications}, 13\penalty0 (1):\penalty0 973, 2022.
\newblock \doi{10.1038/s41467-022-28526-y}.
\newblock URL \url{https://doi.org/10.1038/s41467-022-28526-y}.

\bibitem[Geffner et~al.(2025)Geffner, Didi, Zhang, Reidenbach, Cao, Yim, Geiger, Dallago, Kucukbenli, Vahdat, and Kreis]{proteina}
Tomas Geffner, Kieran Didi, Zuobai Zhang, Danny Reidenbach, Zhonglin Cao, Jason Yim, Mario Geiger, Christian Dallago, Emine Kucukbenli, Arash Vahdat, and Karsten Kreis.
\newblock Proteina: Scaling flow-based protein structure generative models.
\newblock In \emph{Proceedings of the International Conference on Learning Representations (ICLR)}, 2025.

\bibitem[Gerdes et~al.(2023)Gerdes, de~Haan, Rainone, Bondesan, and Cheng]{quantumfield}
Mathis Gerdes, Pim de~Haan, Corrado Rainone, Roberto Bondesan, and Miranda C.~N. Cheng.
\newblock Learning lattice quantum field theories with equivariant continuous flows.
\newblock \emph{SciPost Physics}, 15\penalty0 (6):\penalty0 238, 2023.
\newblock \doi{10.21468/SciPostPhys.15.6.238}.
\newblock URL \url{https://scipost.org/10.21468/SciPostPhys.15.6.238}.

\bibitem[Guan et~al.(2023)Guan, Qian, Peng, Su, Peng, and Ma]{guan20233d}
Jiaqi Guan, Wesley~Wei Qian, Xingang Peng, Yufeng Su, Jian Peng, and Jianzhu Ma.
\newblock 3d equivariant diffusion for target-aware molecule generation and affinity prediction.
\newblock In \emph{Proceedings of the 11th International Conference on Learning Representations (ICLR)}, 2023.

\bibitem[Hassan et~al.(2024)Hassan, Shenoy, Lee, St\"{a}rk, Thaler, and Beaini]{hassan2024flow}
Majdi Hassan, Nikhil Shenoy, Jungyoon Lee, Hannes St\"{a}rk, Stephan Thaler, and Dominique Beaini.
\newblock Et-flow: Equivariant flow-matching for molecular conformer generation.
\newblock In \emph{Advances in Neural Information Processing Systems}, 2024.

\bibitem[Hayes et~al.(2025)Hayes, Rao, Akin, Sofroniew, Oktay, Lin, Verkuil, Tran, Deaton, Wiggert, Badkundri, Shafkat, Gong, Derry, Molina, Thomas, Khan, Mishra, Kim, Bartie, Nemeth, Hsu, Sercu, Candido, and Rives]{hayes2024simulating}
Thomas Hayes, Roshan Rao, Halil Akin, Nicholas~J. Sofroniew, Deniz Oktay, Zeming Lin, Robert Verkuil, Vincent~Q. Tran, Jonathan Deaton, Marius Wiggert, Rohil Badkundri, Irhum Shafkat, Jun Gong, Alexander Derry, Raul~S. Molina, Neil Thomas, Yousuf~A. Khan, Chetan Mishra, Carolyn Kim, Liam~J. Bartie, Matthew Nemeth, Patrick~D. Hsu, Tom Sercu, Salvatore Candido, and Alexander Rives.
\newblock Simulating 500 million years of evolution with a language model.
\newblock \emph{Science}, 373\penalty0 (6557), 2025.

\bibitem[He et~al.(2021)He, Chen, Shen, Dong, Wang, and Lin]{efficientequi}
Lingshen He, Yuxuan Chen, Zhengyang Shen, Yiming Dong, Yisen Wang, and Zhouchen Lin.
\newblock Efficient equivariant network.
\newblock In \emph{Advances in Neural Information Processing Systems 34 (NeurIPS 2021)}, 2021.

\bibitem[Hermann et~al.(2020)Hermann, Sch{\"a}tzle, and No{\'e}]{schrodinger}
Jan Hermann, Zeno Sch{\"a}tzle, and Frank No{\'e}.
\newblock Deep neural network solution of the electronic schrödinger equation.
\newblock \emph{Nature Chemistry}, 12\penalty0 (10):\penalty0 891--897, 2020.

\bibitem[Ho et~al.(2020)Ho, Jain, and Abbeel]{ho2020denoising}
Jonathan Ho, Ajay Jain, and Pieter Abbeel.
\newblock Denoising diffusion probabilistic models.
\newblock \emph{Advances in Neural Information Processing Systems}, 33:\penalty0 6840--6851, 2020.

\bibitem[Hoogeboom et~al.(2022)Hoogeboom, Satorras, Vignac, and Welling]{hoogeboom2022equivariant}
Emiel Hoogeboom, V{\i}ctor~Garcia Satorras, Cl{\'e}ment Vignac, and Max Welling.
\newblock Equivariant diffusion for molecule generation in 3d.
\newblock In \emph{International conference on machine learning}, pp.\  8867--8887. PMLR, 2022.

\bibitem[Huguet et~al.(2024)Huguet, Vuckovic, Fatras, Thibodeau-Laufer, Lemos, Islam, Liu, Rector-Brooks, Akhound-Sadegh, Bronstein, Tong, and Bose]{Huguet2024SequenceAugmentedSMA}
Guillaume Huguet, James Vuckovic, Kilian Fatras, Eric Thibodeau-Laufer, Pablo Lemos, Riashat Islam, Cheng-Hao Liu, Jarrid Rector-Brooks, Tara Akhound-Sadegh, Michael~M. Bronstein, Alexander Tong, and Avishek~Joey Bose.
\newblock Sequence-augmented se(3)-flow matching for conditional protein backbone generation.
\newblock In \emph{Proceedings of the 37th Conference on Neural Information Processing Systems (NeurIPS 2024)}, 2024.

\bibitem[Igashov et~al.(2024)Igashov, St{\"a}rk, Vignac, Schneuing, Satorras, Frossard, Welling, Bronstein, and Correia]{igashov2024equivariant}
Ilia Igashov, Hannes St{\"a}rk, Cl{\'e}ment Vignac, Arne Schneuing, Victor~Garcia Satorras, Pascal Frossard, Max Welling, Michael Bronstein, and Bruno Correia.
\newblock Equivariant 3d-conditional diffusion model for molecular linker design.
\newblock \emph{Nature Machine Intelligence}, pp.\  1--11, 2024.

\bibitem[Ingraham et~al.(2023)Ingraham, Baranov, Costello, Barber, Wang, Ismail, Frappier, Lord, Ng-Thow-Hing, Van~Vlack, Tie, Xue, Cowles, Leung, Rodrigues, Morales-Perez, Ayoub, Green, Puentes, Oplinger, Panwar, Obermeyer, Root, Beam, Poelwijk, and Grigoryan]{Chroma2023}
John~B. Ingraham, Max Baranov, Zak Costello, Karl~W. Barber, Wujie Wang, Ahmed Ismail, Vincent Frappier, Dana~M. Lord, Christopher Ng-Thow-Hing, Erik~R. Van~Vlack, Shan Tie, Vincent Xue, Sarah~C. Cowles, Alan Leung, Jo\~{a}o~V. Rodrigues, Claudio~L. Morales-Perez, Alex~M. Ayoub, Robin Green, Katherine Puentes, Frank Oplinger, Nishant~V. Panwar, Fritz Obermeyer, Adam~R. Root, Andrew~L. Beam, Frank~J. Poelwijk, and Gevorg Grigoryan.
\newblock Illuminating protein space with a programmable generative model.
\newblock \emph{Nature}, 2023.

\bibitem[Jain et~al.(NA)Jain, Ong, Hautier, Chen, Richards, Dacek, Cholia, Gunter, Skinner, Ceder, et~al.]{jainmaterials}
A~Jain, SP~Ong, G~Hautier, W~Chen, WD~Richards, S~Dacek, S~Cholia, D~Gunter, D~Skinner, G~Ceder, et~al.
\newblock The materials project: a materials genome approach to accelerating materials innovation, apl mater. 1 (2013) 011002, NA.

\bibitem[Jain et~al.(2013)Jain, Ong, Hautier, Chen, Richards, Dacek, Cholia, Gunter, Skinner, Ceder, et~al.]{jain2013commentary}
Anubhav Jain, Shyue~Ping Ong, Geoffroy Hautier, Wei Chen, William~Davidson Richards, Stephen Dacek, Shreyas Cholia, Dan Gunter, David Skinner, Gerbrand Ceder, et~al.
\newblock Commentary: The materials project: A materials genome approach to accelerating materials innovation.
\newblock \emph{APL materials}, 1\penalty0 (1), 2013.

\bibitem[Jiao et~al.(2023)Jiao, Huang, Lin, Han, Chen, Lu, and Liu]{jiao2023crystal}
Rui Jiao, Wenbing Huang, Peijia Lin, Jiaqi Han, Pin Chen, Yutong Lu, and Yang Liu.
\newblock Crystal structure prediction by joint equivariant diffusion.
\newblock In \emph{Thirty-seventh Conference on Neural Information Processing Systems}, 2023.
\newblock URL \url{https://openreview.net/forum?id=DNdN26m2Jk}.

\bibitem[Jing et~al.(2022)Jing, Corso, Chang, Barzilay, and Jaakkola]{jing2022torsional}
Bowen Jing, Gabriele Corso, Jeffrey Chang, Regina Barzilay, and Tommi Jaakkola.
\newblock Torsional diffusion for molecular conformer generation.
\newblock In \emph{Advances in Neural Information Processing Systems 35 (NeurIPS 2022)}, pp.\  24240--24253, 2022.
\newblock URL \url{https://papers.nips.cc/paper_files/paper/2022/hash/994545b2308bbbbc97e3e687ea9e464f-Abstract-Conference.html}.

\bibitem[Jing et~al.(2023)Jing, Erives, Pao-Huang, Corso, Berger, and Jaakkola]{PaoHuang2023EigenFoldGPA}
Bowen Jing, Ezra Erives, Peter Pao-Huang, Gabriele Corso, Bonnie Berger, and Tommi Jaakkola.
\newblock Eigenfold: Generative protein structure prediction with diffusion models.
\newblock In \emph{Machine Learning for Drug Discovery Workshop at the International Conference on Learning Representations (ICLR)}, 2023.
\newblock URL \url{https://openreview.net/pdf?id=BgbRVzfQqFp}.

\bibitem[Karras et~al.(2022)Karras, Aittala, Aila, and Laine]{karras2022elucidating}
Tero Karras, Miika Aittala, Timo Aila, and Samuli Laine.
\newblock Elucidating the design space of diffusion-based generative models.
\newblock \emph{Advances in Neural Information Processing Systems}, 35:\penalty0 26565--26577, 2022.

\bibitem[Kim et~al.(2025)Kim, Kim, Kim, Park, and Ahn]{kim2024mofflow}
Nayoung Kim, Seongsu Kim, Minsu Kim, Jinkyoo Park, and Sungsoo Ahn.
\newblock Mofflow: Flow matching for structure prediction of metal-organic frameworks.
\newblock In \emph{Proceedings of the Thirteenth International Conference on Learning Representations (ICLR 2025)}, 2025.
\newblock URL \url{https://openreview.net/forum?id=dNT3abOsLo}.

\bibitem[Kingma et~al.(2021)Kingma, Salimans, Poole, and Ho]{kingma2021variational}
Diederik Kingma, Tim Salimans, Ben Poole, and Jonathan Ho.
\newblock Variational diffusion models.
\newblock \emph{Advances in neural information processing systems}, 34:\penalty0 21696--21707, 2021.

\bibitem[Klein et~al.(2024)Klein, Kr{\"a}mer, and No{\'e}]{klein2024equivariant}
Leon Klein, Andreas Kr{\"a}mer, and Frank No{\'e}.
\newblock Equivariant flow matching.
\newblock \emph{Advances in Neural Information Processing Systems}, 36, 2024.

\bibitem[K{\"{o}}hler et~al.(2020)K{\"{o}}hler, Klein, and No{\'{e}}]{equiflow}
Jonas K{\"{o}}hler, Leon Klein, and Frank No{\'{e}}.
\newblock Equivariant flows: Exact likelihood generative learning for symmetric densities.
\newblock In \emph{Proceedings of the 37th International Conference on Machine Learning, {ICML} 2020, 13-18 July 2020, Virtual Event}, volume 119 of \emph{Proceedings of Machine Learning Research}, pp.\  5361--5370. {PMLR}, 2020.
\newblock URL \url{http://proceedings.mlr.press/v119/kohler20a.html}.

\bibitem[Kozinsky et~al.(2023)Kozinsky, Musaelian, Johansson, and Batzner]{scaledeepequi}
Boris Kozinsky, Albert Musaelian, Anders Johansson, and Simon~L. Batzner.
\newblock Scaling the leading accuracy of deep equivariant models to biomolecular simulations of realistic size.
\newblock In Dorian Arnold, Rosa~M. Badia, and Kathryn~M. Mohror (eds.), \emph{Proceedings of the International Conference for High Performance Computing, Networking, Storage and Analysis, {SC} 2023, Denver, CO, USA, November 12-17, 2023}, pp.\  2:1--2:12. {ACM}, 2023.
\newblock \doi{10.1145/3581784.3627041}.
\newblock URL \url{https://doi.org/10.1145/3581784.3627041}.

\bibitem[Le et~al.(2022)Le, No{\'{e}}, and Clevert]{bioref}
Tuan Le, Frank No{\'{e}}, and Djork{-}Arn{\'{e}} Clevert.
\newblock Representation learning on biomolecular structures using equivariant graph attention.
\newblock In Bastian Rieck and Razvan Pascanu (eds.), \emph{Learning on Graphs Conference, LoG 2022, 9-12 December 2022, Virtual Event}, volume 198 of \emph{Proceedings of Machine Learning Research}, pp.\ ~30. {PMLR}, 2022.
\newblock URL \url{https://proceedings.mlr.press/v198/le22a.html}.

\bibitem[Le et~al.(2024)Le, Cremer, Noé, Clevert, and Schütt]{le2023navigating}
Tuan Le, Julian Cremer, Frank Noé, Djork-Arné Clevert, and Kristof Schütt.
\newblock Navigating the design space of equivariant diffusion-based generative models for de novo 3d molecule generation.
\newblock In \emph{Proceedings of the Twelfth International Conference on Learning Representations (ICLR 2024)}, 2024.

\bibitem[Lin et~al.(2024)Lin, Lee, Zhang, and AlQuraishi]{lin2024genie2}
Yeqing Lin, Minji Lee, Zhao Zhang, and Mohammed AlQuraishi.
\newblock Out of many, one: Designing and scaffolding proteins at the scale of the structural universe with genie 2.
\newblock \emph{arXiv preprint arXiv:2405.15489}, 2024.
\newblock URL \url{https://arxiv.org/abs/2405.15489}.

\bibitem[Lin et~al.(2023)Lin, Akin, Rao, Hie, Zhu, Lu, Smetanin, Verkuil, Kabeli, Shmueli, dos Santos~Costa, Fazel-Zarandi, Sercu, Candido, and Rives]{lin2023evolutionary}
Zeming Lin, Halil Akin, Roshan Rao, Brian Hie, Zhongkai Zhu, Wenting Lu, Nikita Smetanin, Robert Verkuil, Ori Kabeli, Yaniv Shmueli, Allan dos Santos~Costa, Maryam Fazel-Zarandi, Tom Sercu, Salvatore Candido, and Alexander Rives.
\newblock Evolutionary-scale prediction of atomic-level protein structure with a language model.
\newblock \emph{Science}, 2023.

\bibitem[Liu et~al.(2023)Liu, Guo, and Tang]{Liu2022MolecularGPA}
Shengchao Liu, Hongyu Guo, and Jian Tang.
\newblock Molecular geometry pretraining with {SE}(3)-invariant denoising distance matching.
\newblock In \emph{Proceedings of the Eleventh International Conference on Learning Representations (ICLR 2023)}, 2023.
\newblock URL \url{https://openreview.net/forum?id=CjTHVo1dvR}.

\bibitem[Lu et~al.(2022)Lu, Zhou, Bao, Chen, Li, and Zhu]{lu2022dpm}
Cheng Lu, Yuhao Zhou, Fan Bao, Jianfei Chen, Chongxuan Li, and Jun Zhu.
\newblock Dpm-solver: A fast ode solver for diffusion probabilistic model sampling in around 10 steps.
\newblock \emph{Advances in Neural Information Processing Systems}, 35:\penalty0 5775--5787, 2022.

\bibitem[Nichol \& Dhariwal(2021)Nichol and Dhariwal]{nichol2021improved}
Alexander~Quinn Nichol and Prafulla Dhariwal.
\newblock Improved denoising diffusion probabilistic models.
\newblock In \emph{International Conference on Machine Learning}, pp.\  8162--8171. PMLR, 2021.

\bibitem[Oganov \& Glass(2006)Oganov and Glass]{oganov2006crystal}
Artem~R Oganov and Colin~W Glass.
\newblock Crystal structure prediction using ab initio evolutionary techniques: Principles and applications.
\newblock \emph{The Journal of chemical physics}, 124\penalty0 (24), 2006.

\bibitem[Schneuing et~al.(2024)Schneuing, Harris, Du, Didi, Jamasb, Igashov, Du, Gomes, Blundell, Lio, et~al.]{schneuing2024structure}
Arne Schneuing, Charles Harris, Yuanqi Du, Kieran Didi, Arian Jamasb, Ilia Igashov, Weitao Du, Carla Gomes, Tom~L Blundell, Pietro Lio, et~al.
\newblock Structure-based drug design with equivariant diffusion models.
\newblock \emph{Nature Computational Science}, 4\penalty0 (12):\penalty0 899--909, 2024.

\bibitem[Shao et~al.(2021)Shao, Lv, Liu, Shao, Gao, Liu, Wang, and Ma]{Shao2021ASDA}
Xuecheng Shao, Jian Lv, Peng Liu, Sen Shao, P.~Gao, Hanyu Liu, Yanchao Wang, and Yanming Ma.
\newblock A symmetry-orientated divide-and-conquer method for crystal structure prediction.
\newblock \emph{The Journal of chemical physics}, 156 1:\penalty0 014105, 2021.
\newblock URL \url{https://api.semanticscholar.org/CorpusId:238198323}.

\bibitem[Song et~al.(2021{\natexlab{a}})Song, Meng, and Ermon]{song2020denoising}
Jiaming Song, Chenlin Meng, and Stefano Ermon.
\newblock Denoising diffusion implicit models.
\newblock In \emph{Proceedings of the 9th International Conference on Learning Representations (ICLR)}, 2021{\natexlab{a}}.
\newblock URL \url{https://openreview.net/forum?id=St1giarCHLP}.

\bibitem[Song et~al.(2021{\natexlab{b}})Song, Sohl-Dickstein, Kingma, Kumar, Ermon, and Poole]{song2020score}
Yang Song, Jascha Sohl-Dickstein, Diederik~P Kingma, Abhishek Kumar, Stefano Ermon, and Ben Poole.
\newblock Score-based generative modeling through stochastic differential equations.
\newblock In \emph{International Conference on Learning Representations (ICLR)}, 2021{\natexlab{b}}.

\bibitem[Thomas et~al.(2018)Thomas, Smidt, Kearnes, Yang, Li, Kohlhoff, and Riley]{tensorfieldnet}
Nathaniel Thomas, Tess~E. Smidt, Steven Kearnes, Lusann Yang, Li~Li, Kai Kohlhoff, and Patrick Riley.
\newblock Tensor field networks: Rotation- and translation-equivariant neural networks for 3d point clouds.
\newblock In \emph{Advances in Neural Information Processing Systems (NeurIPS) 31}, 2018.

\bibitem[Tong et~al.(2024)Tong, Fatras, Malkin, Huguet, Zhang, Rector-Brooks, Wolf, and Bengio]{tong2023improving}
Alexander Tong, Kilian Fatras, Nikolay Malkin, Guillaume Huguet, Yanlei Zhang, Jarrid Rector-Brooks, Guy Wolf, and Yoshua Bengio.
\newblock Improving and generalizing flow-based generative models with minibatch optimal transport.
\newblock \emph{Transactions on Machine Learning Research}, 2024.

\bibitem[Tong et~al.(2025)Tong, Hoang, Liu, Van~den Broeck, and Niepert]{tong2024learning}
Vinh Tong, Trung-Dung Hoang, Anji Liu, Guy Van~den Broeck, and Mathias Niepert.
\newblock Learning to discretize denoising diffusion odes.
\newblock In \emph{Proceedings of the 13th International Conference on Learning Representations}, 2025.

\bibitem[Vignac et~al.(2023)Vignac, Krawczuk, Siraudin, Wang, Cevher, and Frossard]{digress}
Cl{\'{e}}ment Vignac, Igor Krawczuk, Antoine Siraudin, Bohan Wang, Volkan Cevher, and Pascal Frossard.
\newblock Digress: Discrete denoising diffusion for graph generation.
\newblock In \emph{The Eleventh International Conference on Learning Representations, {ICLR} 2023, Kigali, Rwanda, May 1-5, 2023}, 2023.

\bibitem[Wang et~al.(2024{\natexlab{a}})Wang, Qu, Peng, Wang, Zhu, Chen, and Cao]{wang2024proteus}
Chentong Wang, Yannan Qu, Zhangzhi Peng, Yukai Wang, Hongli Zhu, Dachuan Chen, and Longxing Cao.
\newblock Proteus: exploring protein structure generation for enhanced designability and efficiency.
\newblock \emph{bioRxiv}, pp.\  2024--02, 2024{\natexlab{a}}.

\bibitem[Wang et~al.(2024{\natexlab{b}})Wang, Elhag, Jaitly, Susskind, and Bautista]{Swallow}
Yuyang Wang, Ahmed A.~A. Elhag, Navdeep Jaitly, Joshua~M. Susskind, and Miguel~{\'{A}}ngel Bautista.
\newblock Swallowing the bitter pill: Simplified scalable conformer generation.
\newblock In \emph{Forty-first International Conference on Machine Learning, {ICML} 2024, Vienna, Austria, July 21-27, 2024}, 2024{\natexlab{b}}.

\bibitem[Watson et~al.(2023)Watson, Juergens, Bennett, Trippe, Yim, Eisenach, Ahern, Borst, Ragotte, Milles, Wicky, Hanikel, Pellock, Courbet, Sheffler, Wang, Venkatesh, Sappington, Vázquez~Torres, Lauko, De~Bortoli, Mathieu, Ovchinnikov, Barzilay, Jaakkola, DiMaio, Baek, and Baker]{Watson2023DeNDA}
Joseph~L. Watson, David Juergens, Nathaniel~R. Bennett, Brian~L. Trippe, Jason Yim, Helen~E. Eisenach, Woody Ahern, Andrew~J. Borst, Robert~J. Ragotte, Lukas~F. Milles, Basile I.~M. Wicky, Nikita Hanikel, Samuel~J. Pellock, Alexis Courbet, William Sheffler, Jue Wang, Preetham Venkatesh, Isaac Sappington, Susana Vázquez~Torres, Anna Lauko, Valentin De~Bortoli, Emile Mathieu, Sergey Ovchinnikov, Regina Barzilay, Tommi~S. Jaakkola, Frank DiMaio, Minkyung Baek, and David Baker.
\newblock De novo design of protein structure and function with rfdiffusion.
\newblock \emph{Nature}, 620\penalty0 (7976):\penalty0 1089--1100, 2023.

\bibitem[Wei et~al.(2024)Wei, Omee, Dong, Fu, Song, Siriwardane, Xu, Wolverton, and Hu]{Wei2024CSPBenchABA}
Lai Wei, Sadman~Sadeed Omee, Rongzhi Dong, Nihang Fu, Yuqi Song, E.~Siriwardane, Meiling Xu, Chris Wolverton, and Jianjun Hu.
\newblock Cspbench: a benchmark and critical evaluation of crystal structure prediction.
\newblock In \emph{NA}, 2024.
\newblock URL \url{https://api.semanticscholar.org/CorpusId:270869939}.

\bibitem[Worrall et~al.(2017)Worrall, Garbin, Turmukhambetov, and Brostow]{harmonicnet}
Daniel~E. Worrall, Stephan~J. Garbin, Daniyar Turmukhambetov, and Gabriel~J. Brostow.
\newblock Harmonic networks: Deep translation and rotation equivariance.
\newblock In \emph{2017 {IEEE} Conference on Computer Vision and Pattern Recognition, {CVPR} 2017, Honolulu, HI, USA, July 21-26, 2017}, pp.\  7168--7177. {IEEE} Computer Society, 2017.
\newblock \doi{10.1109/CVPR.2017.758}.
\newblock URL \url{https://doi.org/10.1109/CVPR.2017.758}.

\bibitem[Wu et~al.(2022)Wu, Yang, van~den Berg, Zou, Lu, and Amini]{Wu2022ProteinSGA}
Kevin~E. Wu, Kevin~Kaichuang Yang, Rianne van~den Berg, James Zou, Alex~X. Lu, and Ava~P. Amini.
\newblock Protein structure generation via folding diffusion.
\newblock \emph{Nature Communications}, 15, 2022.
\newblock URL \url{https://api.semanticscholar.org/CorpusId:252668551}.

\bibitem[Xie et~al.(2022)Xie, Fu, Ganea, Barzilay, and Jaakkola]{diffcsp_bl2}
Tian Xie, Xiang Fu, Octavian-Eugen Ganea, Regina Barzilay, and Tommi~S. Jaakkola.
\newblock Crystal diffusion variational autoencoder for periodic material generation.
\newblock In \emph{Proceedings of the 10th International Conference on Learning Representations (ICLR)}, 2022.

\bibitem[Xu et~al.(2022{\natexlab{a}})Xu, Yu, Song, Shi, Ermon, and Tang]{geodiff}
Minkai Xu, Lantao Yu, Yang Song, Chence Shi, Stefano Ermon, and Jian Tang.
\newblock Geodiff: {A} geometric diffusion model for molecular conformation generation.
\newblock In \emph{The Tenth International Conference on Learning Representations, {ICLR} 2022, Virtual Event, April 25-29, 2022}, 2022{\natexlab{a}}.
\newblock URL \url{https://openreview.net/forum?id=PzcvxEMzvQC}.

\bibitem[Xu et~al.(2022{\natexlab{b}})Xu, Yu, Song, Shi, Ermon, and Tang]{xu2022geodiff}
Minkai Xu, Lantao Yu, Yang Song, Chence Shi, Stefano Ermon, and Jian Tang.
\newblock Geodiff: A geometric diffusion model for molecular conformation generation.
\newblock In \emph{International Conference on Learning Representations}, 2022{\natexlab{b}}.
\newblock URL \url{https://openreview.net/forum?id=PzcvxEMzvQC}.

\bibitem[Xu et~al.(2023)Xu, Tong, and Jaakkola]{xu2023stable}
Yilun Xu, Shangyuan Tong, and Tommi Jaakkola.
\newblock Stable target field for reduced variance score estimation in diffusion models.
\newblock \emph{arXiv preprint arXiv:2302.00670}, 2023.

\bibitem[Yim et~al.(2023{\natexlab{a}})Yim, Campbell, Foong, Gastegger, Jim{\'e}nez-Luna, Lewis, Satorras, Veeling, Barzilay, Jaakkola, and No{\'e}]{yim2023fast}
Jason Yim, Andrew Campbell, Andrew Y.~K. Foong, Michael Gastegger, Jos{\'e} Jim{\'e}nez-Luna, Sarah Lewis, Victor~Garcia Satorras, Bastiaan~S. Veeling, Regina Barzilay, Tommi Jaakkola, and Frank No{\'e}.
\newblock Fast protein backbone generation with se(3) flow matching.
\newblock \emph{arXiv preprint arXiv:2310.05297}, 2023{\natexlab{a}}.

\bibitem[Yim et~al.(2023{\natexlab{b}})Yim, Trippe, De~Bortoli, Mathieu, Doucet, Barzilay, and Jaakkola]{yim2023se}
Jason Yim, Brian~L. Trippe, Valentin De~Bortoli, Emile Mathieu, Arnaud Doucet, Regina Barzilay, and Tommi Jaakkola.
\newblock {SE}(3) diffusion model with application to protein backbone generation.
\newblock In \emph{Proceedings of the 40th International Conference on Machine Learning}, 2023{\natexlab{b}}.

\bibitem[Zaverkin et~al.(2024)Zaverkin, Alesiani, Maruyama, Errica, Christiansen, Takamoto, Weber, and Niepert]{zaverkinhigher}
Viktor Zaverkin, Francesco Alesiani, Takashi Maruyama, Federico Errica, Henrik Christiansen, Makoto Takamoto, Nicolas Weber, and Mathias Niepert.
\newblock Higher-rank irreducible cartesian tensors for equivariant message passing.
\newblock In \emph{Conference on Neural Information Processing Systems}, 2024.

\bibitem[Zhang et~al.(2023)Zhang, Qamar, Kang, Jung, Zhang, Bae, and Zhang]{surveygraphdiff}
Mengchun Zhang, Maryam Qamar, Taegoo Kang, Yuna Jung, Chenshuang Zhang, Sung{-}Ho Bae, and Chaoning Zhang.
\newblock A survey on graph diffusion models: Generative {AI} in science for molecule, protein and material.
\newblock \emph{CoRR}, abs/2304.01565, 2023.
\newblock \doi{10.48550/ARXIV.2304.01565}.
\newblock URL \url{https://doi.org/10.48550/arXiv.2304.01565}.

\end{thebibliography}
